\documentclass[10pt,letterpaper]{article}

\usepackage[inline]{enumitem}
\usepackage{placeins}
\usepackage{booktabs}
\usepackage{siunitx}
\usepackage{mathtools}

\usepackage{enumitem}
\usepackage{multirow}
\usepackage{floatrow}
\usepackage[mathscr]{euscript}
\usepackage{float}
\usepackage{hyperref}
\usepackage[utf8]{inputenc}
\usepackage{fourier} 
\usepackage{makecell}
\usepackage{tabularx}
\usepackage{ragged2e}
\usepackage{diagbox}

\usepackage{cite}
\usepackage{amsmath,amssymb,amsfonts}
\usepackage{pifont}
\usepackage{cleveref}

\usepackage{algorithm}
\usepackage{algpseudocode}

\newcommand{\IndentAlgoMulti}[1]{
\newcount\foo
\foo=#1
\loop
\hskip\algorithmicindent
  \advance \foo -1
\ifnum \foo>0
\repeat
}

\usepackage[top=0.85in,left=2.75in,footskip=0.75in]{geometry}

\usepackage{amsmath,amssymb}

\usepackage{changepage}

\usepackage{textcomp,marvosym}
\usepackage[notextcomp]{stix}

\usepackage{cite}

\usepackage{nameref,hyperref}

\usepackage[right]{lineno}

\usepackage{microtype}
\DisableLigatures[f]{encoding = *, family = * }

\usepackage[table]{xcolor}

\usepackage{array}

\newcolumntype{+}{!{\vrule width 2pt}}

\newlength\savedwidth

\raggedright
\usepackage[aboveskip=1pt,labelfont=bf,labelsep=period,justification=raggedright,singlelinecheck=off]{caption}

\makeatletter
\renewcommand{\@biblabel}[1]{\quad#1.}
\makeatother

\usepackage{lastpage,fancyhdr,graphicx}
\usepackage{epstopdf}
\fancyheadoffset[L]{2.25in}
\fancyfootoffset[L]{2.25in}
\begin{document}
\vspace*{0.2in}

\begin{flushleft}
{\Large
\textbf\newline{SCGG: A Deep Structure-Conditioned Graph Generative Model} 
}
\newline
\\
Faezeh Faez\textsuperscript{1},
Negin Hashemi Dijujin\textsuperscript{1},
Mahdieh Soleymani Baghshah\textsuperscript{1*},
Hamid R. Rabiee\textsuperscript{1*}
\\
\bigskip
\textbf{1} Department of Computer Engineering, Sharif University of Technology, Tehran, Iran
\\
\bigskip

%
%





* Rabiee@sharif.edu (HRR), Soleymani@sharif.edu (MS)

\end{flushleft}
\section*{Abstract}
\textcolor{black}{Deep learning-based graph generation approaches have remarkable capacities for graph data modeling, allowing them to solve a wide range of real-world problems. Making these methods able to consider different conditions during the generation procedure even increases their effectiveness by empowering them to generate new graph samples that meet the desired criteria. This paper presents a conditional deep graph generation method called SCGG that considers a particular type of structural conditions. Specifically, our proposed SCGG model takes an initial subgraph and autoregressively generates new nodes and their corresponding edges on top of the given conditioning substructure. The architecture of SCGG consists of a graph representation learning network and an autoregressive generative model, which is trained end-to-end. Using this model, we can address graph completion, a rampant and inherently difficult problem of recovering missing nodes and their associated edges of partially observed graphs. Experimental results on both synthetic and real-world datasets demonstrate the superiority of our method compared with state-of-the-art baselines.}

\section{Introduction}
 \textcolor{black}{With the ever-increasing growth of data collection and production technologies, large amounts of data is readily accessible. In many cases, some kind of relationship exists between data entities, which, if taken into consideration, can lead to more precise data analyses. Such relationships are mostly represented by graph data structures, and that is why graph-related research has become a widely discussed topic in many areas including chemistry\cite{mahmood2021masked}, medical applications\cite{ghorbani2021gkd}, social network studies\cite{min2021stgsn}, and knowledge graph-related research \cite{chen2021dacha}. Most recent studies are dedicated to graph representation learning\cite{chen2022structure, han2022geometric}, aiming to obtain suitable representations of nodes, edges or the entire graph in continuous space to be further utilized by downstream tasks.}\par
 \textcolor{black}{Graph generation is another important branch of graph-related research, which often benefits from the results of graph representation learning studies. This research field has a history of several decades. It has recently been revived by receiving renewed attention from scholars, mainly due to the advances in machine learning, and in particular deep learning techniques. The goal of graph generation is to provide models that can generate new graph samples from the desired data distributions. Thus, similar to generative methods in other data domains such as image\cite{deng2022gram}, text\cite{liu2021dexperts}, and speech\cite{chen2022epg2s}, graph generative approaches can bring substantial capacity for graph data modeling to address various real-world problems such as drug design \cite{li2018multi}, understanding and modeling the interactions in social networks \cite{grover2019graphite}, and human diseases diagnosis \cite{yang2022collaborative}.}
 \par
\textcolor{black}{One of the desired and essential properties of generative methods is their ability to carry out the generation procedure in a controlled manner so that the produced samples comply with predetermined conditions by having the required characteristics. In this regard, numerous studies have been conducted to develop conditional generative models in different data domains, such as image\cite{kang2020contragan} and text\cite{guo2021conditional}. Initial steps\cite{simonovsky2018graphvae, yang2019conditional, lim2020scaffold, jin2020hierarchical, ommi2022ccgg} have also been taken to make graph generators conditional, however, compared to the work performed in other data domains and also compared to the needs and capacities of this field, much remains to be done.}\par
\textcolor{black}{In addition to what we have discussed so far, there is a common problem manifesting itself when working with different types of data. Specifically, in many cases, the data is not completely available, which can be caused due to various reasons such as limitations of data collection tools, issues related to privacy, or inadequacy of storage space. This can significantly degrade the performance of data analysis methods. Therefore, it is often crucial to recover the missing part of the data before processing it; hence, various methods have been proposed in different data domains to address this challenge. Regarding the graph data, many methods have also been developed for years \cite{zhou2009predicting, liu2010link} to predict missing links between graph nodes, and researchers are still seriously pursuing a solution for it \cite{teji2022predicting}. However, an intrinsically more complicated challenge arises when the graph nodes are missing. We will refer to this problem as graph completion, which, unlike the widely investigated problem of link prediction, has been much less addressed despite its importance and pervasiveness.}\par
\textcolor{black}{To address the issues mentioned above, we propose \textbf{Structure-Conditioned Graph Generator (SCGG)}, an end-to-end deep learning-based conditional graph generative approach. The SCGG model takes an initial subgraph as the structural condition. It then autoregressively performs the graph generation procedure by adding new nodes and predicting the inter-links between the new nodes and those in the conditioning subgraph, as well as the intra-links between the new nodes themselves. In this way, our generative model ensures the existence of desired subgraphs in final generated graphs, which can have several applications in both molecular and non-molecular domains. Specifically, for designing molecular graphs, the existence of desired chemical substructures can bring certain chemical properties to the final molecules. Moreover, regarding the non-molecular graphs, the SCGG model can be best utilized to solve the graph completion problem in which some graph nodes and their corresponding edges are totally missing. Our study focuses on the latter application, but the proposed SCGG model can be easily extended to be used in molecular applications as well. In this regard, a partially observed graph is given to the model as a structural condition. Then the generated nodes by the model and their associated edges will be treated as the recovered missing nodes and the edges connecting them to each other, as well as to the partially observed graph nodes.}\par
\textcolor{black}{In summary, we present the following contributions in this work:}
\begin{itemize}
    \item \textcolor{black}{We introduce SCGG, a conditional graph generation approach, which autoregressively generates graphs based on a given structural condition.} 
    \item \textcolor{black}{The architecture of our SCGG model consists of a graph representation learning network and a recurrent neural network (RNN), where the former is mainly used to take into account the structural condition, and the latter captures the generation history.}
    \item \textcolor{black}{We use our proposed SCGG model to address the graph completion problem to benefit from the power and potential of a deep generative model for solving an inherently difficult and complex problem, which as a result has been relatively less investigated so far.} \textcolor{black}{To the extent of our knowledge, this is the first time that a completely deep learning-based model is designed in such a way that it can specifically tackle this problem.}
    \item \textcolor{black}{We conduct extensive experiments on both synthetic and real-world datasets to compare the performance of our proposed model against the baselines. The experimental results indicate that the SCGG model outperforms the state-of-the-art baselines in terms of the distance between the generated graphs and the ground-truth ones.} 
\end{itemize}\par
\textcolor{black}{The rest of the paper is organized as follows. In Section \ref{sec:related_work}, we review the previous work related to our research. In Section \ref{sec:notations}, we introduce the notations used in the paper and define the problem. In Section \ref{sec:scgg}, we explain our proposed SCGG model in detail. Experimental details and results are discussed in Section \ref{sec:experiments}. Finally, in Section \ref{sec:conclusion} we conclude the paper.} 

\section{Related work}\label{sec:related_work}
\textcolor{black}{In line with what we discussed earlier, our proposed SCGG model is a structure-based conditional graph generation approach that one of its main applications is graph completion. Therefore, in the following, we review the literature in two related areas.}
\subsection{Graph generation}
\textcolor{black}{Graph generation is a field of research seeking to generate new graph structures with certain characteristics, which dates back to several decades ago, and is still a hot topic for research. In contrast to the early methods\cite{erdHos1960evolution, watts1998collective, holland1983stochastic, albert2002statistical}, which relied on manually-designed procedures to construct graphs with predetermined statistical properties, the more recent ones are data-driven, utilizing the available graph samples in datasets to train models that can more effectively generate new graphs. The latter approaches typically employ different deep learning techniques and generation strategies, and accordingly, they can be classified into several categories \cite{faez2021deep}. The autoregressive approaches, which adopt step-by-step strategies for generating graphs, are the most relevant methods to our research. DeepGMG \cite{li2018learning} is an example of them proposing a repetitive decision-making process to generate graphs gradually. GraphRNN \cite{you2018graphrnn} is among the well-known and influential approaches, which first maps each graph into a sequence of nodes and then processes one node per time step using RNNs to model the distribution of the resulting sequences. The method has inspired a number of subsequent approaches like MolecularRNN \cite{popova2019molecularrnn}, which extends GraphRNN to generate molecular graphs with specific chemical features. Bacciu et al. \cite{bacciu2020edge}, GraphGen \cite{goyal2020graphgen}, and GHRNN \cite{xianduo2022hierarchical}, on the other hand, convert graphs to sequences of edges instead of nodes, and then go through distribution modeling with RNNs. Besides, there are some other autoregressive methods that utilize the attention mechanism to empower their generative models. Regarding this, GRAN \cite{liao2019efficient} proposes to add a block of new nodes in each step, and to compute the representations of the graph nodes, it employs an attentive message passing mechanism.}\par 
\textcolor{black}{In addition to the methods mentioned above that are more related to our proposed approach, there are other categories of modern graph generation approaches, the most noteworthy of which are autoencoder-based methods \cite{simonovsky2018graphvae, guo2020interpretable, li2020dirichlet, du2021deep, du2022disentangled, du2022interpretable}, RL-based approaches \cite{you2018graph, ahn2020guiding, darvariu2021goal}, GAN-based generating strategies \cite{de2018molgan, yang2019conditional, yang2022collaborative}, and flow-based models \cite{shi2020graphaf, luo2021graphdf}.} \par 
\textcolor{black}{A key point to notice is that regardless of what category these methods fall into and what techniques they employ to solve the problem, an important capability of them is to consider desired conditions during generation so that the resulting graphs meet the expected characteristics. Hence, the problem of conditional graph generation arises. In this regard, GraphVAE \cite{simonovsky2018graphvae} conditions both the encoder and the decoder of its VAE on a label vector for the molecular graph generation. CONDGEN \cite{yang2019conditional} adopts a similar approach (i.e., concatenating a condition vector to the VAE latent variable) to incline the model towards generating graphs with desired characteristics. Lim et al. \cite{lim2020scaffold} and HierVAE \cite{jin2020hierarchical} guarantee the existence of intended chemical substructures in the output molecular graphs. CCGG \cite{ommi2022ccgg} makes the GRAN \cite{liao2019efficient} model class-conditional, allowing it to generate graphs of desired classes. However, despite the efforts that have gone into conditional graph generation, there is still a vital need to develop more and more approaches that can capture various types of conditions. In this regard, the SCGG model is a generative method designed to handle special conditions, which are of structural type.}
\subsection{Graph completion}
\textcolor{black}{In many cases, a part of a graph structure is unavailable for various reasons. Hence, it is necessary to reconstruct the missing information  prior to further processing. Most of the methods developed for this purpose try to perform link prediction \cite{jiao2021temporal, wang2021self, nayyeri2021trans4e}, although a more complicated problem arises when the graph nodes are missing. Therefore, due to the complexity of addressing this problem, which we refer to as graph completion, so far few methods have been presented to solve it. Regarding this, KronEM\cite{kim2011network} utilizes a combination of the Expectation-Maximization framework and the Kronecker graphs model to infer the missing nodes and their corresponding edges. SAMI \cite{sina2015sami} adopts a clustering approach for solving the missing node problem by heavily relying on the existence of missing node indicators, which are often unattainable in real scenarios. Masrour et al. \cite{masrour2015network} and JCSL \cite{rafailidis2016network} utilize side information about the graph nodes to perform network completion; however, this information may not be accessible in all cases. More recently, DeepNC \cite{9229516} was introduced, which first learns the likelihood of the data by training the GraphRNN \cite{you2018graphrnn} model. It then uncovers the missing parts of a graph by proposing a greedy optimization algorithm, aiming to maximize the obtained likelihood. Although DeepNC is an innovative approach that has obtained satisfactory results, it is not learning-based, so it cannot directly learn from the data for the specific task of graph completion. However, our proposed method trains an end-to-end model to address this problem. Furthermore, unlike some graph completion methods mentioned above, the SCGG does not depend on the existence of side information, which may not be reachable in many situations.}

\section{Notations and problem definition}\label{sec:notations}
\textcolor{black}{In this section, we define the notations used in the paper and present the problem definition. For convenience, we summarize the notations in Table \ref{tab:notations}.}\par
\begin{table}
\caption{\label{tab:notations} \textbf{Notations in this paper.}}
\centering
\textcolor{black}{
\begin{tabular}{p{2 cm} p{5.50cm}} \toprule
    \textbf{Notation} & \textbf{Description} \\ \midrule
    $G_0$   & {An initial graph.}     \\ \midrule
    $V_0$   & {The node set of $G_0$.}     \\ \midrule
    $E_0$   & {The edge set of $G_0$.}     \\ \midrule
    {n}  & {The number of nodes in $G_0$, n=$|V_0|$.} \\ \midrule
    {$\pi_n$}  & {An ordering of $G_0$ nodes.} \\ \midrule
    {$N_i^{\pi_n}$}  & {The sequence representing how the $i$-th node of $G_0$ under the ordering $\pi_n$ connects to $G_0$'s nodes.}\\ \midrule
    $G$   & {The graph that contains an initial graph $G_0$ as a subgraph.}\\ \midrule
    $V$   & {The node set of $G$.}\\ \midrule
    $E$   & {The edge set of $G$.}\\\midrule
    $\pmb{\mathcal{G}}$   & {The random variable associated with graph structures.}\\ \midrule
    $\tilde{V}$   & {The set of new nodes added to $G_0$ to form the graph $G$.}\\ \midrule
    $\tilde{E}$   & {The set of edges connecting the new nodes to each other, as well as to those nodes in $G_0$.}\\ \midrule
    $m$   & {The number of new nodes, m=$|\tilde{V}|$.}\\ \midrule
    $\pi_m$   & {An ordering of the new nodes $\tilde{V}$.}\\\midrule
    $\tilde{N_i}^{\pi_n}$   & {The sequence representing the links connecting the $i$-th node of $G_0$ under the ordering $\pi_n$ to the new nodes ordered by $\pi_m$.}\\\midrule
    $\tilde{M}_j^{\pi_m} $   & {The sequence representing the links between the $j$-th new node and each of the new nodes under the ordering $\pi_m$.}\\\midrule
    $S_0^{\pi_n}$   & {The notational abbreviation for $\{N_1^{\pi_n}, \cdots, N_n^{\pi_n}\}$.}\\\midrule
    $\tilde{S}^{\pi_n, \pi_m}$   & {The notational abbreviation for $\{\tilde{N_1}^{\pi_n}, \cdots, \tilde{N_n}^{\pi_n}, \tilde{M}_1^{\pi_m}, \cdots, \tilde{M}_m^{\pi_m}\}$}\\\midrule
    $G^\prime$   & {The graph induced from $G$ by removing the intra-connections between the set of new nodes}\\ \bottomrule
\end{tabular}}
\end{table}

\textcolor{black}{We denote an initial graph as $G_0=(V_0, E_0)$, where $V_0$ and $E_0$ are the node and the edge sets, respectively, and $|V_0|=n$. 
Under an ordering $\pi_n$ of these $n$ nodes, we represent the $i$-th node's links by the following sequence:
\begin{equation}
    N_i^{\pi_n} = \big(x_k\big)_{k = 1}^n, \ x_k \in \{0, 1\}
\end{equation}
where $x_k$ takes value of 1 if the $i$-th node is connected to the $k$-th node and $0$ otherwise.}\par
\textcolor{black}{Considering $G_0$ as the structural condition, the objective of our research is to learn to sample from the conditional probability distribution $P(\pmb{\mathcal{G}}|G_0)$  in order to generate graph $G=(V, E)$, which includes $G_0$ as a subgraph, i.e., $V_0 \subset V$ and $E_0 \subset E$. This can be done by first adding the node set $\tilde{V}$, with $|\tilde{V}| = m$ and $\tilde{V} = V - V_0$.} 
\textcolor{black}{Then, to connect new nodes, the edge set $\tilde{E}$ will be generated, where $\tilde{E}=E-E_0$. More specifically, $\tilde{E}$ consists of: 
\begin{enumerate*}[itemjoin={,\quad}]
\item the inter-connections between new nodes and those in $G_0$
\item the intra-connections between the new nodes themselves.
\end{enumerate*}
To represent the inter-connections between new nodes and the $i$-th node of $G_0$ under the ordering $\pi_n$, we use the below sequence:
\begin{equation}
    \tilde{N_i}^{\pi_n} = \big(x_l\big)_{l = 1}^m, \ x_l \in \{0, 1\}
\end{equation}
where we consider a node ordering $\pi_m$ of the $m$ new nodes, and $x_l$ is 1 if the $i$-th node of $G_0$ has a link to the $l$-th new node and 0 otherwise.
Moreover, regarding the intra-connections, we denote the $j$-th new node's connections to the nodes in $\tilde{V}$ by the following sequence:
\begin{equation}
    \tilde{M}_j^{\pi_m} = \big(x_p\big)_{p = 1}^m, \ x_p \in \{0, 1\}
\end{equation}
where similarly to the previous formulas, $x_p$ takes the value of 1 if there is a link connecting the $j$-th and the $p$-th new nodes (under the ordering $\pi_m$) and 0 otherwise.}

\section{SCGG: Structure-Conditioned Graph Generator}\label{sec:scgg}
\textcolor{black}{We approach a specific type of structure-conditioned graph generation that takes an initial substructure and starts to generate new nodes and their associated edges on top of the given conditioning substructure. To this end, we propose the SCGG model, whose architecture is composed of a graph representation learning network and an autoregressive generative model, which is trained in an end-to-end manner.}
\textcolor{black}{In this section, we present the details of the SCGG model. In this regard, we first elucidate the problem formulation and the model architecture. Next, we describe the procedure employed to prepare the data for model training. Then, we discuss the training and inference phases and elaborate on the implementation details.}
\subsection{Formulation}
\textcolor{black}{As mentioned in the Section \ref{sec:notations}, in this work we intend to learn to sample from the distribution $P(\pmb{\mathcal{G}}|G_0)$ to conditionally generate the graph $G$ given an arbitrary initial graph $G_0$. To do so, our SCGG model first estimates this conditional probability distribution and then samples from the resulting estimated distribution. As it is not easy to work directly in the graph space, we reformulate the problem to deal with the following distribution:
    \begin{equation}\label{eq:cond_prob}
    P(\tilde{S}^{\pi_n, \pi_m}|S_0^{\pi_n}) = 
    P(\tilde{N_1}^{\pi_n}, \cdots, \tilde{N_n}^{\pi_n}, \tilde{M}_1^{\pi_m}, \cdots, \tilde{M}_m^{\pi_m}|N_1^{\pi_n}, \cdots, N_n^{\pi_n})
    \end{equation}
    where $S_0^{\pi_n}$ and $\tilde{S}^{\pi_n, \pi_m}$ are the notational abbreviations for $\{N_1^{\pi_n}, \cdots, N_n^{\pi_n}\}$ and $\{\tilde{N_1}^{\pi_n}, \cdots, \tilde{N_n}^{\pi_n}, \tilde{M}_1^{\pi_m}, \cdots, \tilde{M}_m^{\pi_m}\}$, respectively, and the new problem formulation relates to the original one through the below equation:
    \begin{equation}\label{eq:marginal}
        P(\pmb{\mathcal{G}}|G_0) = \sum_{\pi_n, \pi_m} P(\tilde{S}^{\pi_n, \pi_m}|S_0^{\pi_n})
    \end{equation}}
    \par
    \textcolor{black}{To further decompose the probability in Eq. \ref{eq:cond_prob}, we follow the chain rule and therefore this conditional probability can be rewritten as follows:
    \begin{equation}\label{eq:2}
        \begin{split}
        P(\tilde{S}^{\pi_n, \pi_m}|S_0^{\pi_n}) =& \\&
        P(\tilde{N_1}^{\pi_n}|N_1^{\pi_n}, \cdots, N_n^{\pi_n}) \times\\&
        P(\tilde{N_2}^{\pi_n}|N_1^{\pi_n}, \cdots, N_n^{\pi_n}, \tilde{N_1}^{\pi_n}) \times\\&
        \;\;\vdots \\&
        P(\tilde{N_n}^{\pi_n}|N_1^{\pi_n}, \cdots, N_n^{\pi_n}, \tilde{N_1}^{\pi_n}, \cdots, \tilde{N}_{n-1}^{\pi_n}) \times\\&
        P(\tilde{M}_1^{\pi_m}|N_1^{\pi_n}, \cdots, N_n^{\pi_n}, \tilde{N_1}^{\pi_n}, \cdots, \tilde{N}_n^{\pi_n}) \times\\&
        P(\tilde{M}_2^{\pi_m}|N_1^{\pi_n}, \cdots, N_n^{\pi_n}, \tilde{N_1}^{\pi_n}, \cdots, \tilde{N}_n^{\pi_n}, \tilde{M}_1^{\pi_m}) \times\\&
        \;\;\vdots \\&
        P(\tilde{M}_m^{\pi_m}|N_1^{\pi_n}, \cdots, N_n^{\pi_n}, \tilde{N_1}^{\pi_n}, \cdots., \tilde{N}_n^{\pi_n}, \tilde{M}_1^{\pi_m}, \cdots, \tilde{M}_{m-1}^{\pi_m})
        \end{split}
    \end{equation}}
\par
\textcolor{black}{Our proposed SCGG method trains a novel network architecture in an end-to-end manner to model the complex distribution in Eq. \ref{eq:2}.}

\subsection{Model architecture}
\textcolor{black}{The model architecture of SCGG consists of two main components, namely, a graph representation learning network and an autoregressive generative model (i.e., an RNN). In the following, we explain these components in detail and discuss the role each plays in the task of structure-conditioned graph generation.} 
\subsubsection{Graph Feature Learning Network}
\begin{figure*}[h]
\begin{adjustwidth}{-2.25in}{0in}
\centering
\includegraphics[width=1.07\textwidth]{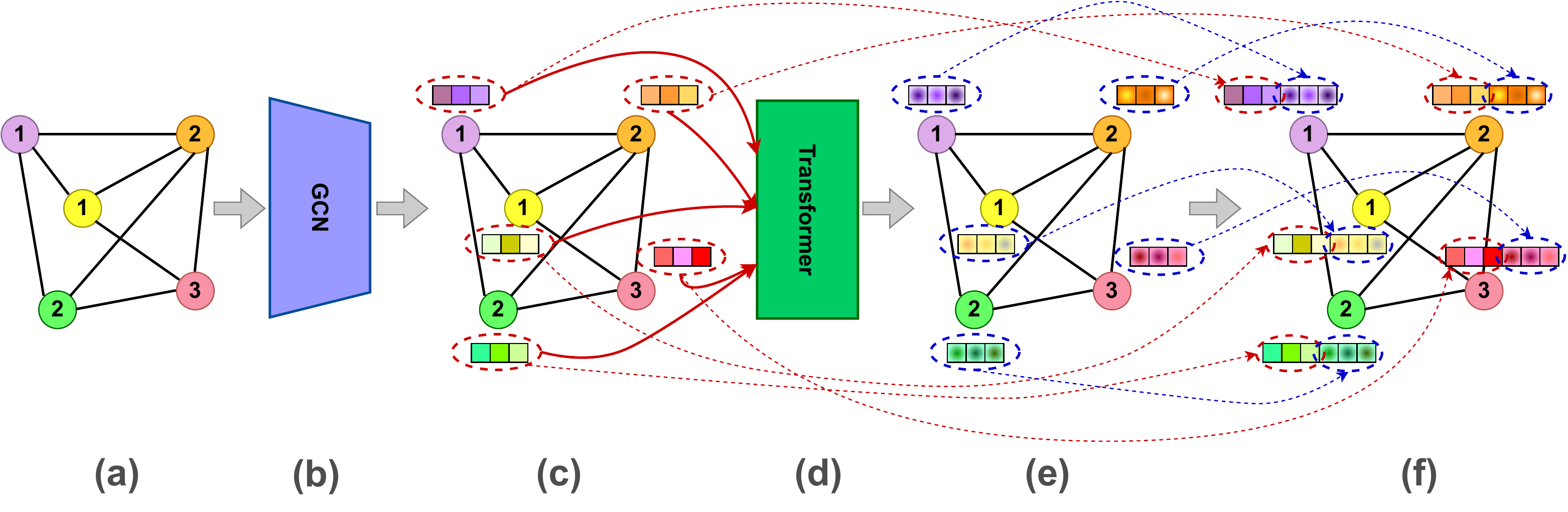}
\vspace*{-0.23cm}
\centering
\caption[]{\textcolor{black}{An illustration of the Graph Feature Learning Network and its workflow.}
\begin{enumerate*}[label=(\alph*)]
\item \textcolor{black}{An input graph. }
\item \textcolor{black}{The Graph Convolutional Network.}
\item \textcolor{black}{Continuous representations learned for graph nodes by the GCN.}
\item \textcolor{black}{The Transformer network that takes the node embeddings computed by the GCN as input and outputs new contextualized features of graph nodes.} 
\item \textcolor{black}{The node features learned by the Transformer network (shown using small squares colored with radial gradients).}
\item \textcolor{black}{The final representations of graph nodes acquired by concatenating the embeddings computed by the GCN and the Transformer network. Here, dashed arrows are drawn to easily track what sub-features a final node feature consists of.}
\end{enumerate*}}
\label{GFLN}
\end{adjustwidth}
\end{figure*}

\textcolor{black}{The SCGG method needs appropriate representations of graph nodes beforehand to perform distribution modeling. Therefore, it utilizes a graph representation learning network that employs both a graph convolutional network (GCN) and a Transformer network to learn meaningful node features. Below, we give a brief background of GCNs and Transformers. Furthermore, we elaborate on how each of them contributes to obtaining the final nodes' features in our model.} 
\begin{itemize}
    \item \textbf{Graph Convolutional Network (GCN)}\\
    \textcolor{black}{It is often difficult to directly work in the complex and discrete graph space. Therefore, in many cases, obtaining continuous representations of nodes, edges, or the whole graph is necessary prior to any upcoming tasks. Employing Graph Convolutional Networks addresses this problem. The main idea of GCNs originates from the fact that a node's representation can be obtained by taking into account the features of its own and its neighbors. This is because the neighbors in a graph (i.e., directly or indirectly connected nodes) usually share some common characteristics and information.\\
    Formally, the layer-wise propagation rule of GCNs can be generally formulated as below:
 \begin{equation}
     X^{l+1}=\phi(AX^lW^l)
 \end{equation}
 where $X^l \in \mathbb{R}^{N\times D_l}$ is the nodes' feature matrix at the $l$-th GCN layer, $N$ is the number of graph nodes, $D_l$ is the number of features obtained for a node by the previous GCN layer, and $X^0$ is set to be the initial feature matrix given as input to the GCN; $A \in \mathbb{R}^{N\times N}$ is the adjacency matrix \cite{dudziak2020brp, niu2019multi} or a variant of it \cite{kipf2016semi, zhang2018end}; $W^l \in \mathbb{R}^{D_l\times D_{l+1}}$ is the learnable parameter matrix of the $l$-th GCN layer, which maps $D_l$ feature channels to $D_{l+1}$ channels; $\phi$ is a non-linear activation function; $X^{l+1} \in \mathbb{R}^{N\times D_{l+1}}$ is the output feature matrix produced by the $l$-th GCN layer.\\
Considering this background, our proposed Graph Feature Learning Network first applies $L$ layers of GCN to the input graph. This way, a continuous representation is computed for each graph node based on its neighbors' information.}
    \item \textbf{Transformer network}\\
    \textcolor{black}{In this work, we intend to autoregressively model the distribution in Eq. \ref{eq:cond_prob}, which is conditioned on $\{N_1^{\pi_n}, \cdots, N_n^{\pi_n}\}$. We do so by feeding the representations of graph nodes one at a time into the RNN. Thus, in order to perform conditional distribution modeling in this way, it is necessary to learn rich node representations so that all the graph nodes can make their own contribution to compute each node's embedding. In other words, we need the representation of a node not only to contain the information of its close neighbors, but also to include the information of relatively distant nodes that share some similar characteristics with it. However, an $L$-layer GCN only considers information in $L$-hop neighborhoods to obtain node representations, even if there are some dependencies between farther nodes. Therefore, our proposed Graph Feature Learning Network utilizes a Transformer encoder, which has shown promising results in contextualized representation learning. The following gives a quick overview of its architecture and workflow.\\
    According to \cite{vaswani2017attention}, the Transformer encoder layer consists of a multi-head attention block and a feedforward network, each followed by a residual addition and a layer normalization. A multi-head attention block consists of multiple attention heads, each working in a separate subspace to compute new contextualized representations corresponding to different aspects of dependencies between data entities. To be more precise, each attention head takes as input $X^L \in \mathbb{R}^{N \times D_L}$ (in our case it is the feature matrix computed for graph's nodes by applying $L$ layers of GCN) and projects it into three matrices $Q = X^LW_q\in \mathbb{R}^{N \times D_k}$, $K = X^LW_k\in \mathbb{R}^{N \times D_k}$, and $V = X^LW_v \in \mathbb{R}^{N \times D_v}$ (i.e., query, key, and value, respectively), where $W_q$, $W_k\in \mathbb{R}^{D_L\times D_k}$ and $W_v\in \mathbb{R}^{D_L\times D_v}$ are learnable matrices. Then, the attention scores for each query are computed over the rows of the value matrix $V$ by performing an inner product of that query and all the key matrix $K$ rows. By doing so, a new contextualized representation is calculated for each query as a weighted summation of the value matrix rows.}
\end{itemize}
\textcolor{black}{Considering these remarks regarding the GCN and the Transformer, the final nodes' representations are obtained via concatenating the features computed by each of the two networks. Fig. \ref{GFLN} shows an overview of the proposed Graph Feature Learning Network.}
\subsubsection{Autoregressive generative model}
\textcolor{black}{As mentioned earlier, we want to model the conditional distribution in Eq. \ref{eq:cond_prob}. To do so, we decompose it as the product of $n+m$ conditional distributions in Eq. \ref{eq:2}, and then go through modeling them. Each condition in Eq. \ref{eq:2} can be divided into two parts: 
\begin{enumerate*}[label=(\alph*)]
\item $\{N_1^{\pi_n}, \cdots, N_n^{\pi_n}\}$ that is the initial structural condition regarding to $G_0$ and
\item The remaining part of the condition derived by applying the chain rule, which relates to the generation history.
\end{enumerate*} The former is primarily captured by our Graph Feature Learning Network, and the latter is handled using an autoregressive generative model, namely an RNN. More specifically, the embeddings obtained by the Graph Feature Learning Network are fed into the RNN one at a time, and the RNN proceeds. This way, the RNN keeps the generation history such that at each step, the corresponding hidden state maintains the information of the graph generated until that time.}

\subsection{Data preparation}
\begin{figure*}[h]
\begin{adjustwidth}{-0.25in}{0in}
\centering
\includegraphics[width=1\textwidth]{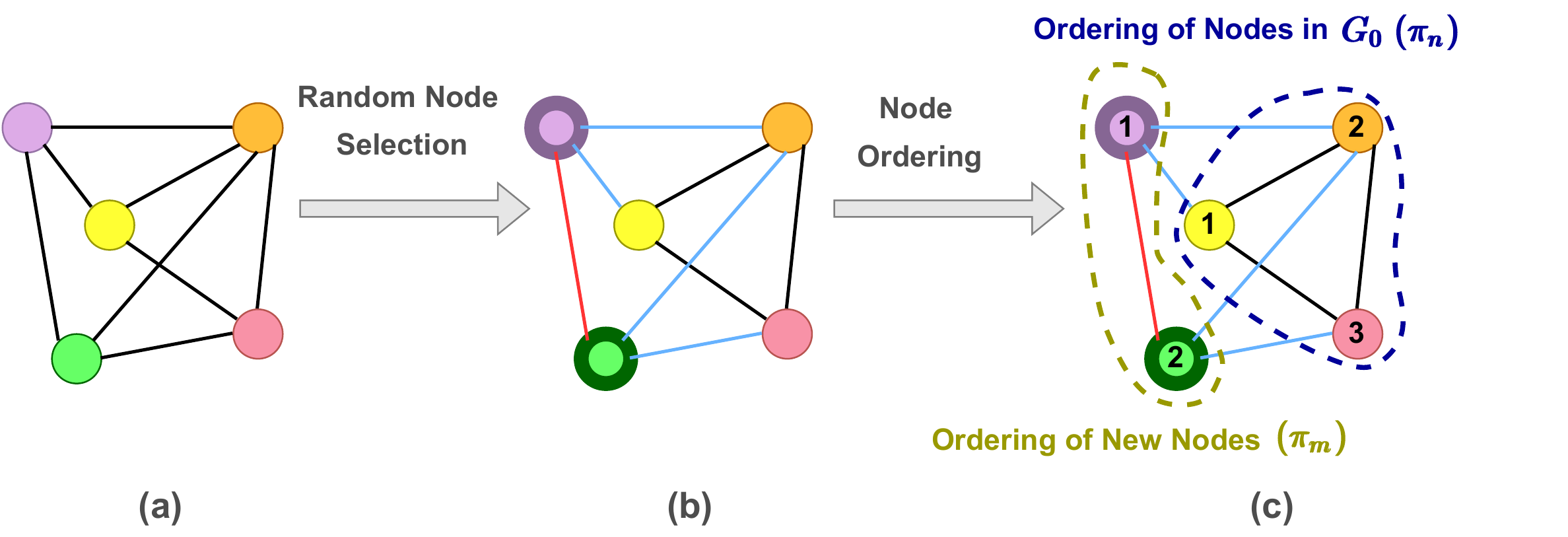}
\vspace*{-0.23cm}
\centering
\caption[]{\textcolor{black}{An illustration of the procedure of preparing the training data.}
\begin{enumerate*}[label=(\alph*)]
\item \textcolor{black}{An input graph.}
\item \textcolor{black}{A number of $m$ nodes are selected at random to be further treated as the new nodes. In this picture, $m=2$ and the selected nodes (i.e., the green and the purple ones) are shown with thick borders. Furthermore, the inter-connections between new nodes and those in $G_0$ are depicted by blue lines, and the only intra-connection between the new nodes is shown using a red line.} 
\item \textcolor{black}{An ordering $\pi_n$ is applied to the nodes in $G_0$. Moreover, another node ordering, denoted by $\pi_m$, is applied to the new nodes.}
\end{enumerate*}}
\label{data_preprocessing}
\end{adjustwidth}
\end{figure*}

\textcolor{black}{Making the data suitable as an input to our SCGG model is a prerequisite for training. Therefore, we perform a data preparation procedure before feeding it to the model. This procedure includes determining the set of new nodes $\tilde{V}$, identifying the resulting initial graph $G_0$, and applying orderings on these two sets of nodes.
An example of the data preparation procedure before model training is illustrated in Fig. \ref{data_preprocessing}. First, $m$ nodes are randomly selected from the main graph $G$ to form the set of new nodes. Therefore, the $n$ unselected nodes and those edges connecting them to each other are further treated as the initial graph $G_0$. The reason behind this random node selection is that each subset of $n$ nodes (i.e., the unselected ones) from the original graph has the chance to contribute to the model training as an initial graph. Thus, the model gains the ability to perform structure-conditioned graph generation given an arbitrary graph $G_0$ at test time. Afterwards, orderings are applied to the nodes such that the initial graph nodes are ordered by $\pi_n$, and the new nodes follow the order specified by $\pi_m$.}

\subsection{Training}
\textcolor{black}{To train the SCGG model, we first give it two versions of each graph $G$. The first version corresponds to the initial graph $G_0$. The second version, which we denote by $G^\prime$, is obtained by removing the intra-connections between pairs of nodes belonging to $\tilde{V}$. The Graph Feature Learning Network takes these two graphs as inputs and separately calculates nodes' representations for each of them, as formulated below:
\begin{equation}
    R = [r_1, r_2, \cdots, r_n] = f_{emb}(G_0)
\end{equation}
\begin{equation}
    R^\prime = [r^\prime_{1}, r^\prime_{2}, \cdots, r^\prime_{n}, r^\prime_{n+1}, \cdots, r^\prime_{n+m}] = f_{emb}(G^\prime)
\end{equation}Next, a subset of the computed representations are fed into the RNN one by one in the order specified by $\pi_n$ and $\pi_m$. More precisely, the RNN first takes the representations of $G_0$'s nodes computed based on the first version of the graph. Then, it receives as input the representations of the new nodes obtained by feeding the $G^\prime$ into the Graph Feature Learning Network. To put it another way, the final representations to be fed into the RNN are as follows:
\begin{equation}
    R^{\prime \prime} = \big[r^{\prime \prime}_i\big]_{i = 1}^{n+m} = [r_1, r_2, \cdots, r_n, r^\prime_{n+1}, \cdots, r^\prime_{n+m}]
\end{equation}
The reason for this is that at test time, we only have access to an initial graph $G_0$ knowing nothing about how the set of new nodes are connected to each other as well as to the rest of the graph, but as the RNN proceeds, it predicts the inter-connections between the new nodes and the nodes of $G_0$. Thus, when the RNN finishes processing the last node of $G_0$, all inter-connections have been predicted and $G^\prime$ can be constructed on top of $G_0$. At this point, it is time to complete the graph structure by predicting the intra-links between the new nodes. This requires that we have a proper representation for each new node, which can be obtained based on the most complete available version of the graph structure, i.e., the $G^\prime$.}\par 
\textcolor{black}{Moreover, each cell of the RNN takes as its second input the ground truth labels of the previous cell. Therefore, the input for the $i$-th RNN cell is obtained as follows:
\begin{equation}
    x_i = Concat(r^{\prime \prime}_i, s_{i-1})
\end{equation}
where $r^{\prime \prime}_i$ is the representation of the $i$-th node and $s_{i-1}\in \mathbb{R}^{m}$ is the vector of ground truth labels determining whether the $i-1$-th node has links to each of the new nodes or not. Next, by considering both the current input $x_i$ and the previous hidden state $h_{i-1}$, the RNN outputs probabilities regarding the link existence between the current node and each new node. This is done using two functions $f_{RNN}$ and $f_{out}$ according to the following formulations:
\begin{equation}
    h_i = f_{RNN}(x_i, h_{i-1})
\end{equation}
\begin{equation}
    \phi_i = f_{out}(h_i)
\end{equation}
where $\phi_i\in \mathbb{R}^{m}$ is the $i$-th step probabilistic output. Furthermore, the step loss $L_i$ is a binary cross entropy (BCE) between the predicted outputs and the ground truth labels, which is formulated in the below equation:
\begin{equation}
    \text{BCE}(\phi_i, s_i) = 
    -\frac{1}{m}\sum_{k=1}^m\Big(s_i[k]\log\phi_i[k] + (1-s_i[k])\log(1-\phi_i[k])\Big)
\end{equation}}\par
\textcolor{black}{The whole network, including the Graph Feature Learning Network and the RNN, is trained in an end-to-end manner. Algorithm \ref{alg:train} summarizes the training procedure of our SCGG model.}

\begin{figure}[t]
\centering
\includegraphics[width=0.5\textwidth]{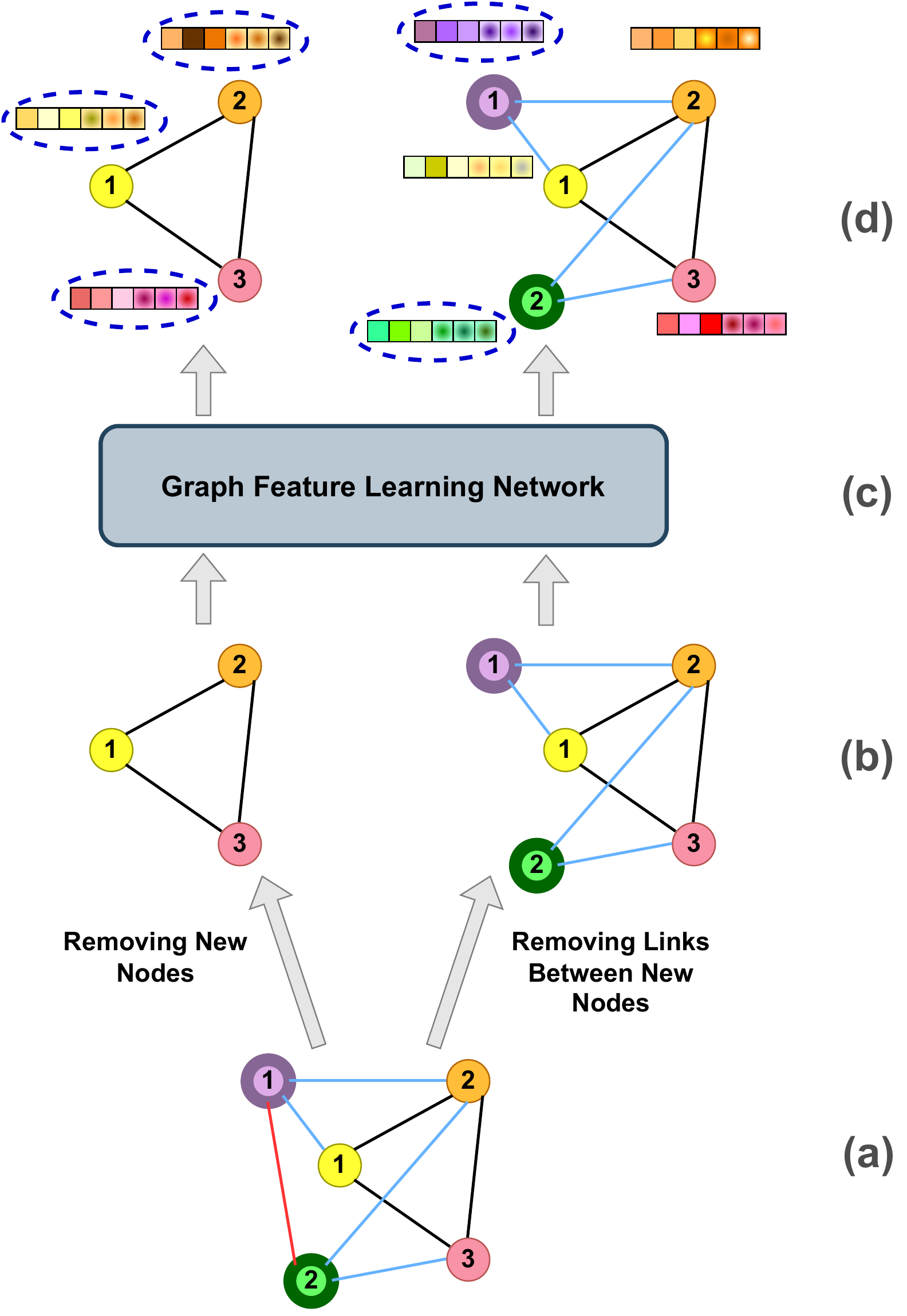}
\vspace*{-0.5cm}
\begin{center}
\caption[]{\textcolor{black}{An overview of the workflow employed to obtain the required nodes' features in the training phase.}
\begin{enumerate*}[label=(\alph*)]
\item \textcolor{black}{An input training graph after applying the preparation procedure shown in Fig. \ref{data_preprocessing}}
\item \textcolor{black}{Two versions are made from the main graph. The one on the left will be treated as the initial graph (i.e., the $G_0$), and the graph on the right, which we denote in the paper by $G^\prime$, is obtained from the original graph by removing the intra-connection between the new nodes, i.e., the red link.}
\item \textcolor{black}{The Graph Feature Learning Network, whose architecture is illustrated in detail in Fig. \ref{GFLN}.}
\item \textcolor{black}{The features computed for each node of the graphs. The ones around which blue dashed ovals are drawn will be further used by the RNN.}\label{feature_learning:d}
\end{enumerate*}}
\label{feature_learning}
\end{center}
\end{figure}

\begin{figure*}[h]
\begin{adjustwidth}{-2.25in}{0in}
\centering
\includegraphics[width=1.02\textwidth]{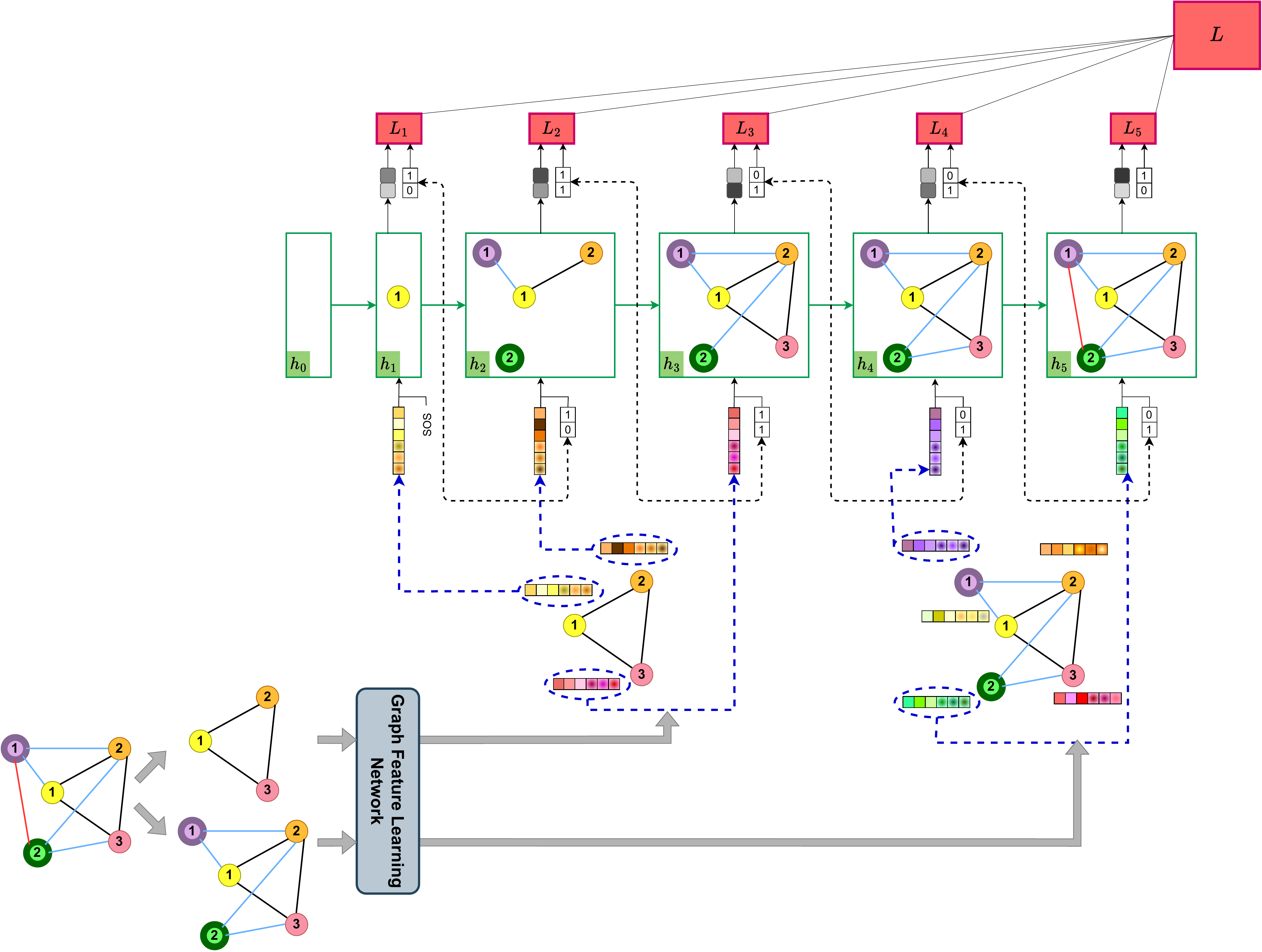}
\vspace*{-0.2cm}
\centering\caption[]{\textcolor{black}{An example of the SCGG model at training time. For each graph node, including those in the initial graph (i.e., the $G_0$) and the ones in the set of new nodes (i.e., the $\tilde{V}$), the model outputs a probability distribution of link existence between that node and each new node (the probabilistic outputs are depicted by grey squares, and the darker the colors, the higher the probabilities). To do this, at each step, a recurrent unit takes the features computed for one of the graph nodes (see Fig. \ref{feature_learning}), as well as the previous node's true connections and the hidden state of the previous recurrent unit. In this regard, the nodes of $G_0$ (ordered by $\pi_n$) are first fed into the model, which are then followed by the new nodes (ordered by $\pi_m$). Thus, the model learns to first generate the inter-links between the new nodes and those of $G_0$, and then predict the intra-links between the new nodes. The parameters of both the Graph Feature Learning Network and the RNN are updated by minimizing the total loss $L$ that is obtained via aggregating the step losses $L_i$.} }
\label{train}
\end{adjustwidth}
\end{figure*}

\begin{algorithm}
\caption{\textcolor{black}{Training Algorithm of SCGG Model}}\label{alg:train}
\begin{algorithmic}[1]
\Require \textcolor{black}{Dataset of training graphs $\mathcal{D}$, number of new nodes $m$}
\Ensure \textcolor{black}{Learned functions $f_{emb}$, $f_{RNN}$, and $f_{out}$}
\For{\textcolor{black}{$\forall G \in \mathcal{D}$}}
\State\textcolor{black}{Build $G_0$ and $G^\prime$ from the graph $G$}
\EndFor
\For{\textcolor{black}{number of training iterations}}
\For{\textcolor{black}{$\forall G \in \mathcal{D}$}}
\State \textcolor{black}{$R = [r_1, r_2, \cdots, r_n] = f_{emb}(G_0)$}
\State \textcolor{black}{$R^\prime = [r^\prime_{1}, r^\prime_{2}, \cdots, r^\prime_{n}, r^\prime_{n+1}, \cdots, r^\prime_{n+m}] = f_{emb}(G^\prime)$}
\State \textcolor{black}{$R^{\prime \prime} = \big[r^{\prime \prime}_i\big]_{i = 1}^{n+m} = [r_1, r_2, \cdots, r_n, r^\prime_{n+1}, \cdots, r^\prime_{n+m}]$}
\State \textcolor{black}{$s_0$ = \texttt{sos}; Initialize $h_0$; $\textit{L} = 0$}
\For{\textcolor{black}{$i$ from $1$ to $n+m$}}
\State \textcolor{black}{$x_i = Concat(r^{\prime \prime}_i, s_{i-1})$}
\State \textcolor{black}{$h_i = f_{RNN}(x_i, h_{i-1})$}
\State \textcolor{black}{$\phi_i = f_{out}(h_i)$}
\State \textcolor{black}{$\textit{L}$ = $\textit{L}$ + BCE($\phi_i$, $s_i$)}
\EndFor
\State \textcolor{black}{$\textit{L}$ = $\textit{L}$ / ($n+m$)}
\State \textcolor{black}{Update model parameters by performing backpropagation to minimize the loss func-} 
\Statex \IndentAlgoMulti{2} \textcolor{black}{ tion $\textit{L}$}
\EndFor
\EndFor
\end{algorithmic}
\end{algorithm}
\textcolor{black}{An example showing the SCGG model at training time is presented in Figures \ref{feature_learning} and \ref{train}, where the graph of Fig. \ref{data_preprocessing} is used as training data. First, the representations of the nodes in both $G_0$ and $G^\prime$ are computed by the Graph Feature Learning Network, which is illustrated in Fig. \ref{feature_learning}. Then, the obtained representations for the $G_0$'s nodes (see the left half of Fig. \ref{feature_learning} \labelcref{feature_learning:d}) are given to the RNN in the order specified by $\pi_n$. Accordingly, as depicted in Fig. \ref{train}, in the first RNN step, it is the turn of node 1 (indicated by a yellow circle) to be processed, and thus its features are passed on to the first recurrent unit. The network then estimates the conditional probability distribution $P(\tilde{N_1}^{\pi_n}|N_1^{\pi_n}, N_2^{\pi_n}, N_3^{\pi_n})$, i.e., the probability of connecting the yellow node to each of the new nodes (the green and the purple ones). Afterwards, the step loss is calculated by taking the network output and the true labels (the first label is 1 because the yellow and the purple nodes are connected, and the second label is 0 as there is no edge between the yellow and the green nodes). In the second step, the second node's features (indicated by orange color) along with the true labels of the previous (yellow) node and the previous hidden state are given to the recurrent cell. Then the network outputs an estimation of the $P(\tilde{N_2}^{\pi_n}|N_1^{\pi_n}, N_2^{\pi_n}, N_3^{\pi_n}, \tilde{N_1}^{\pi_n})$. The same procedure continues until all nodes of $G$, including the ones in $G_0$ and the set of new nodes (i.e., $\tilde{V}$), are fed into the network. Thus, in the third step, the network outputs the probability of $P(\tilde{N_3}^{\pi_n}|N_1^{\pi_n}, N_2^{\pi_n}, N_3^{\pi_n}, \tilde{N_1}^{\pi_n},  \tilde{N_2}^{\pi_n})$ by taking into account the features of the third (pink) node in graph $G_0$. In the subsequent step, when all the initial graph's nodes have been processed, it is time to go through the new nodes in the order specified by $\pi_m$. Thus, the features computed for the first new node (displayed in purple color in the right half of Fig. \ref{feature_learning} \labelcref{feature_learning:d}) is given to the RNN to generate the probability $P(\tilde{M}_1^{\pi_m}|N_1^{\pi_n}, N_2^{\pi_n}, N_3^{\pi_n}, \tilde{N_1}^{\pi_n}, \tilde{N_2}^{\pi_n}, \tilde{N}_3^{\pi_n})$. Next, in the fifth step, the second new node's features (indicated in green) are fed into the recurrent network to produce the probability distribution $P(\tilde{M}_2^{\pi_m}|N_1^{\pi_n}, N_2^{\pi_n}, N_3^{\pi_n}, \tilde{N_1}^{\pi_n}, \tilde{N_2}^{\pi_n}, \tilde{N}_3^{\pi_n}, \tilde{M}_1^{\pi_m})$.}\par
\textcolor{black}{In order to elaborate a bit more on Fig. \ref{train}, it is worth mentioning that each step's hidden state contains the information of a subgraph of the main graph (i.e., $G$). This subgraph includes the already processed graph nodes and the links connecting them to each other as well as their connections to each of the new nodes. It also includes links between the current node and the previous ones. For example, in Fig. \ref{train}, in the third training step, two nodes (i.e., the yellow and the orange ones) have been processed and the pink node's features are fed into the recurrent unit as part of its input. Hence, the hidden state $h_3$ maintains a subgraph containing the link between the yellow and the orange nodes as well as the links between these nodes and the new nodes (shown by blue lines). It also retains the links between the current (pink) node and both the yellow and the orange ones that have been fed into the network in the first two steps.} 

\subsection{Inference}
\begin{figure*}[h]
\begin{adjustwidth}{-2.25in}{0in}
\centering
\includegraphics[width=1.07\textwidth]{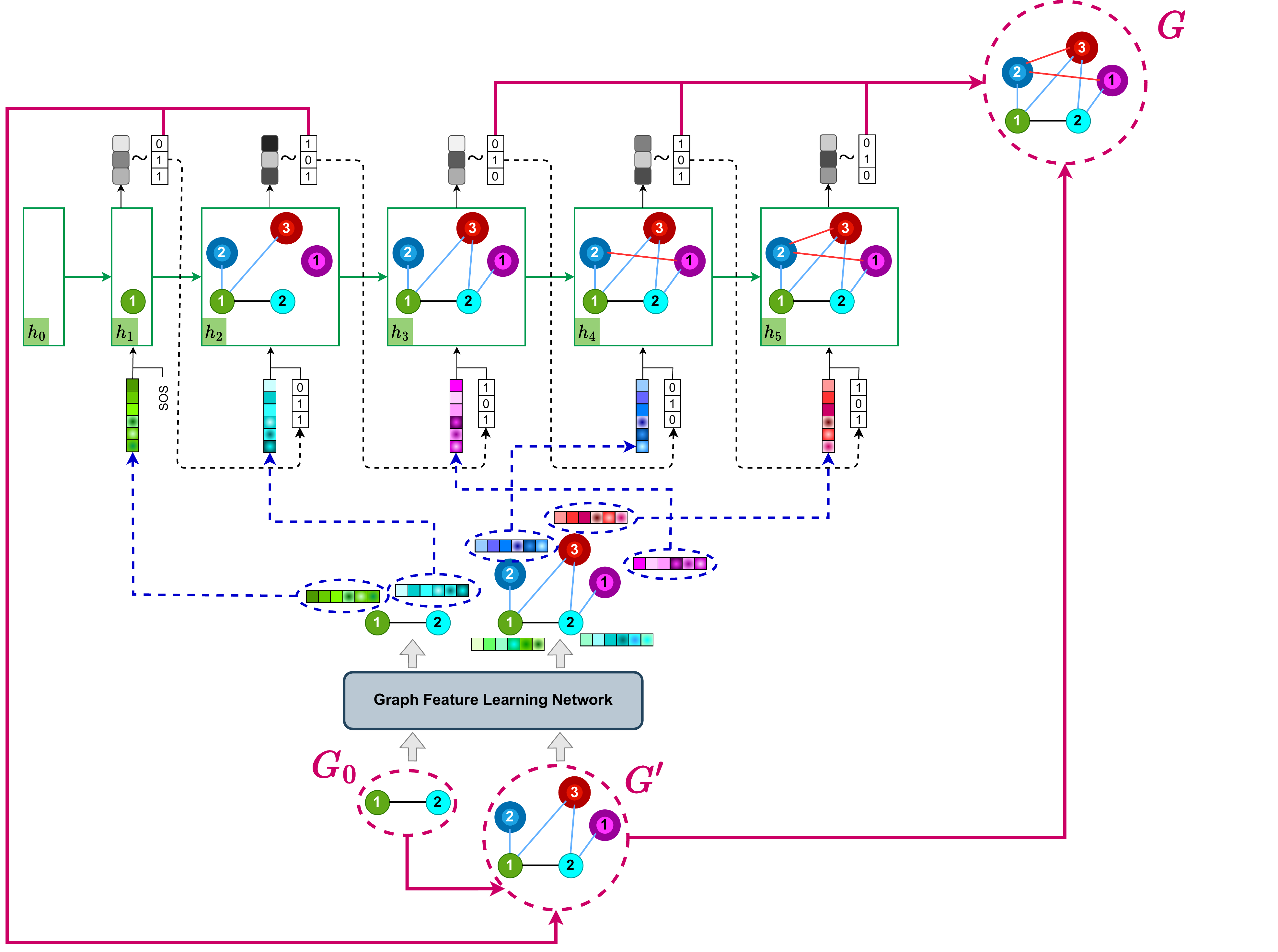}
\vspace*{-0.2cm}
\caption{\textcolor{black}{An example illustrating the SCGG model at inference time. In this example, $m=3$ and a graph $G_0$ consisting of two nodes is given to the model as the structural condition. At first, the Graph Feature Learning Network computes representations for the $G_0$'s nodes, which are then used as part of the RNN input. Next, the RNN proceeds for two steps and outputs the probabilities of the inter-connections between these two nodes and each of the new nodes. Therefore, all the inter-links are generated by sampling from the produced probabilities. At this point, it is time to construct graph $G^\prime$ based on the $G_0$ and the generated links. Next, $G^\prime$ is passed into the Graph Feature Learning Network to calculate the representations of its nodes. In this step, the representations of the new nodes are given to the RNN one by one in order to generate the intra-connections. Finally, the graph $G$ is constructed on top of the $G^\prime$ by considering the generated intra-links.}}
\label{fig:inference}
\end{adjustwidth}
\end{figure*}

\textcolor{black}{In the inference stage, an initial graph $G_0$ is given as the structural condition. Then, using the learned functions $f_{emb}$, $f_{RNN}$, and $f_{out}$, the model starts generating graph $G$ by adding new nodes to $G_0$ and predicting the inter-links between the new nodes and those of $G_0$, as well as the intra-links between the new nodes themselves. Algorithm \ref{alg:inference} describes the steps of the SCGG model at inference time. Moreover, Fig. \ref{fig:inference} illustrates the inference workflow of the SCGG by a toy example.}  
\begin{algorithm}
\caption{\textcolor{black}{Inference Algorithm of SCGG Model}}\label{alg:inference}
\begin{algorithmic}[1]
\Require \textcolor{black}{$f_{emb}$, $f_{RNN}$, $f_{out}$, $m$, $G_0$}
\Ensure \textcolor{black}{$G$}
\State \textcolor{black}{$R = [r_1, r_2, \cdots, r_n] = f_{emb}(G_0)$}
\State \textcolor{black}{$s_0$ = \texttt{sos}; Initialize $h_0$}
\For{\textcolor{black}{$i$ from $1$ to $n$}}
\State \textcolor{black}{$x_i = Concat(r_i, s_{i-1})$}
\State \textcolor{black}{$h_i = f_{RNN}(x_i, h_{i-1})$}
\State \textcolor{black}{$\phi_i = f_{out}(h_i)$}
\State \textcolor{black}{$s_i \sim \phi_i$} \Comment{\textcolor{black}{Sample the inter-connections between the $i$-th node of $G_0$ and the set of new}}
 \Statex \IndentAlgoMulti{3} \textcolor{black}{nodes}
 \EndFor
 \State \textcolor{black}{Construct graph $G^\prime$ on top of $G_0$ using the sampled links [$s_1, s_2, \cdots, s_n$]}
 \State \textcolor{black}{$R^\prime = [r^\prime_{1}, r^\prime_{2}, \cdots, r^\prime_{n}, r^\prime_{n+1}, \cdots, r^\prime_{n+m}] = f_{emb}(G^\prime)$}
 \For{\textcolor{black}{$j$ from $n+1$ to $n+m$}}
 \State \textcolor{black}{$x_j = Concat(r^\prime_j, s_{j-1})$}
\State \textcolor{black}{$h_j = f_{RNN}(x_j, h_{j-1})$}
\State \textcolor{black}{$\phi_j = f_{out}(h_j)$}
\State \textcolor{black}{$s_j \sim \phi_j$} \Comment{\textcolor{black}{Sample the intra-connections between the $j-n$-th new node and each of the}}
\Statex \IndentAlgoMulti{4} \textcolor{black}{new nodes}
 \EndFor
 \State \textcolor{black}{Construct graph $G$ on top of $G^\prime$ using the sampled links [$s_{n+1}, s_{n+2}, \cdots, s_{n+m}$]}
\end{algorithmic}
\end{algorithm}
\subsection{Implementation details}\label{subsec:impl}
\textcolor{black}{The proposed model is implemented using the PyTorch Library \cite{paszke2017automatic}. As previously discussed, the function $f_{emb}$ consists of a graph convolutional network (GCN) and a Transformer network. In this regard, we use a two-layer GCN with the embedding size of each layer equaling 16. ReLU activation followed by a batch normalization layer are used between the two GCN layers. Besides, our Transformer has one encoder layer with 8 attention heads and a dropout of 0.1. We use 4 layers of GRU cells with a 128-dimensional hidden state to implement the function $f_{RNN}$. For the function $f_{out}$, a two-layer multilayer perceptron (MLP) is employed with 64 hidden units in the middle and a ReLU nonlinearity between the layers. Further, the Adam optimizer is used with the learning rate of 0.003, and the model is trained for 100 epochs with a minibatch size of 32. Moreover, for the choice of $\pi_n$ and $\pi_m$, we use uniform random orderings to maximize an approximation of the marginal likelihood in Eq. \ref{eq:marginal}, which becomes intractable to compute exactly as the size of graphs increases.}
\section{Experiments}\label{sec:experiments}
\textcolor{black}{In this section, we first elaborate on both the synthetic and the real-world datasets we used for evaluation purpose. Then, we outline the state-of-the-art baselines with which we compare our SCGG model. Next, the evaluation metric is explained, followed by describing the experimental setup. Finally, we discuss the results of our proposed approach, as well as the ones of the competitor methods.}
\subsection{Datasets}
\begin{table}[!ht]
\begin{adjustwidth}{-2.25in}{0in}
\begin{center}
\textcolor{black}{
\begin{tabular}[t]{|lm{0.15\textwidth}m{0.15\textwidth}m{0.15\textwidth}m{0.15\textwidth}m{0.15\textwidth}m{0.1\textwidth}|}
\hline
Dataset      & Min. \# Nodes                                   & Max. \# Nodes &     Avg. \# Nodes      &     Std. \# Nodes      &     Avg. Sparsity      &  \# Graphs      \\ \hline \hline
Grid          & 25 & 361 & 144 & 75.83 & 0.97 & 225      \\ \hline
IMDBBINARY          & 12 & 136 & 19.77 & 10.06 & 0.51 & 1000       \\ \hline
IMDBMULTI          & 11 & 89 & 18.83 & 9.75 & 0.44 & 686      \\ \hline
Enzymes     & 15 & 125 & 34.61 & 13.86 & 0.87 & 545     \\ \hline
NCI1      & 11 & 111 & 30.05 & 13.48 & 0.92 & 4075     \\ \hline
Protein         & 100 & 500 & 257.87 & 105.50 & 0.98 & 918      \\ \hline
\end{tabular}}
\caption{\label{tab:dataset} \textbf{Statistics of datasets used in the experiments.}}
\end{center}
\end{adjustwidth}
\end{table}
\textcolor{black}{We evaluate the performance of our proposed method on a variety of synthetic and real-world datasets. In the following, we provide a brief description of each dataset. Moreover, Table \ref{tab:dataset} summarizes the key statistics of them.}\\
\begin{itemize}
\item \textbf{Grid:} \textcolor{black}{It is a synthetic dataset consisting of standard 2D grid graphs. }\\
\item \textbf{IMDBBINARY:} \textcolor{black}{This dataset consists of ego-networks derived from actor/actress collaborations based on the information of movies belonging to the Action and Romance genres on IMDB. For each graph, nodes represent actors/actresses, and if a pair of them appears in the same movie, a link connects their corresponding nodes in the graph.}\\
\item \textbf{IMDBMULTI:} \textcolor{black}{The same explanation given for the IMDBBINARY dataset is valid for this dataset as well, except that the movies belong to the Comedy, Romance, and Sci-Fi genres.}\\
\item \textbf{Enzymes:} \textcolor{black}{This dataset consists of graphs each representing a protein tertiary structure from the BRENDA enzyme database \cite{schomburg2004brenda}. More precisely, a graph's nodes represent secondary structure elements (SSEs) and an edge connects two nodes if their corresponding SSEs are neighbors along the amino acid sequence or one of the three nearest neighbors in space.}\\
\item \textbf{NCI1:} \textcolor{black}{It is a biological graph dataset published by the National Cancer Institute (NCI). Each graph in the dataset represents a chemical compound screened for its activity against the growth of human tumors.}\\
\item \textbf{Protein:} \textcolor{black}{This dataset contains protein graphs \cite{dobson2003distinguishing}. Each graph represents a protein with nodes corresponding to amino acids. If the distance between two amino acids of a protein is less than 6 Angstroms, their corresponding nodes are connected in the graph.}
\end{itemize}
\subsection{State-of-the-art approaches}
\textcolor{black}{We compare our approach with several well-known state-of-the-art methods, explanations of which are provided in the following.}
\begin{itemize}
\item \textbf{KronEM \cite{kim2011network}. }\textcolor{black}{This is an old and well-known network completion method that combines the Expectation-Maximization (EM) framework with the Kronecker graphs model \cite{leskovec2010kronecker} to infer missing nodes and their corresponding edges in partially observed graphs. To do this, in each EM iteration, the method first utilizes the observed part of a graph to estimate model parameters (the M-step), and then it infers the missing part of that graph using the estimated model (the E-step).}
\item \textbf{GraphRNN-S \cite{you2018graphrnn}.} \textcolor{black}{This is a very famous autoregressive deep graph generator that first transforms graphs into sequences and then models the corresponding data distribution using RNNs. At each step, the method adds a new node to the currently generated graph and predicts the links connecting it to the previous nodes. Aside from that, GraphRNN-S makes a simplistic assumption that a node's links are independent of each other, and therefore models them by a multi-layer perceptron.}
\item \textbf{GraphRNN \cite{you2018graphrnn}.} \textcolor{black}{This is the full GraphRNN model, which is relatively similar to GraphRNN-S, with the difference that it does not take into account the edge independence simplifying assumption. Therefore, to capture the interdependencies between a node's edges, it employs another recurrent neural network called the edge-level RNN.}
\item \textbf{DeepNC \cite{9229516}.} \textcolor{black}{This is the most recent graph completion baseline that utilizes a deep generative model of graphs, namely GraphRNN-S, to infer the missing parts of a partially observable network. To this end, the method first learns a likelihood over data by training the GraphRNN-S model. Then, it proposes a sequence of algorithmic steps to recover the network in a greedy fashion, trying to maximize the learned likelihood. The fact needed to be noted here is that although this method uses the probabilities generated by a deep generative model of graphs to make algorithmic decisions, it is not considered a totally deep learning-based approach. However, if a model is specifically trained to address the problem of graph completion, it can achieve higher performance.}
\item \textbf{EvoGraph \cite{park2018evograph}.} \textcolor{black}{This is a graph upscaling method, which expands an initial input graph $G_0=(V_0, E_0)$ in $K$ stages by adding $|E_0|$ new edges at each stage. The method considers a set of candidate new nodes in every expansion phase, and adds each new edge by choosing one of its endpoints from the current nodes and the other from the candidate ones. In order to provide a fair comparison between EvoGraph and other methods, we make a slight change to its upscaling process by terminating it right after the insertion of the $m$-th new node.}
\end{itemize}
\subsection{Evaluation metric}
\textcolor{black}{Similar to \cite{9229516}, we use \textit{Graph Edit Distance (GED)} \cite{sanfeliu1983distance} as the evaluation metric to assess the performance of our SCGG method and the baselines. In this regard, if we denote a generated or completed graph by $\hat{G}$ and its corresponding ground truth graph by $G$, the GED between these two graphs, which shows how dissimilar they are, can be formulized as follows:
\begin{equation}
    d(\hat{G}, G) = \min_{\lambda \in \gamma(\hat{G}, G)} \sum_{e_i \in \lambda} c(e_i) \label{ged}
\end{equation}
where $\gamma(\hat{G}, G)$ is the set of all edit paths converting $\hat{G}$ to a graph that is isomorphic to $G$. Moreover, $c(e_i)$ is the cost of an edit operation $e_i$, which in the same way as \cite{9229516}, we set it to 1 for all operations. Additionally, as with \cite{9229516}, we normalize the GED computed for each pair of graphs by the average of their sizes.}\par
\textcolor{black}{Along with our brief overview of GED, one important point to note is that enumerating all the discussed edit paths requires employing a combinatorial search procedure with exponential time complexity, and therefore the exact solution to this problem is NP-complete\cite{zeng2009comparing}. Hence, we utilize an approximation approach \cite{fischer2017improved} for computing GED scores.}
\subsection{Experimental setup}
\textcolor{black}{In addition to what we have explained in Section \ref{subsec:impl} concerning the details of implementing our SCGG model, in this subsection, we elaborate on the remainder of the details regarding the experimental setup. In this respect, to train our model, we select a random subset of 80\% of the graphs in each dataset. A similar approach is also followed to train other learning-based baselines (i.e., GraphRNN-S and GraphRNN). We then make use of the remaining 20\% of graphs for model testing. More specifically, for each graph $G$ in the test set, we perform the following two steps for 10 iterations:}
\begin{itemize}
    \item \textcolor{black}{We randomly choose a number of $m$ nodes from the original test graph $G$ and remove these nodes and their associated edges to acquire a subgraph $G_0$.}
    \item \textcolor{black}{We then feed the obtained subgraph to all the competing methods and compare their results to the ground truth graph $G$.} 
\end{itemize}
\textcolor{black}{Afterwards, for each graph in the test data, we average the GED scores calculated in 10 iterations and compute their standard deviation. Finally, for each value of the parameter $m$, we report the average of the GED scores, as well as the average of standard deviations computed over the whole test set.}
\subsection{Results and discussion}
\begin{table}[!ht]
\begin{adjustwidth}{-2.25in}{0in}
\begin{center}
\small
\textcolor{black}{
\scalebox{0.95}{
{\renewcommand{\arraystretch}{2} 
\begin{tabular}[t]{|l|c|c|c|c|c|c|c|}
\hline 
\multicolumn{2}{|c|}{\diagbox{Method}{Dataset}} & Grid                                   & IMDBBINARY &     IMDBMULTI      &     Enzymes      &     NCI1      &  Protein     \\ \hline
\multicolumn{2}{|l|}{KronEM ($Y_1$)}         & 0.5875 $\pm$ 0.1900 & 0.5634 $\pm$ 0.2065 & 0.5650 $\pm$ 0.2068 & 0.5698 $\pm$ 0.1952 & 0.6521 $\pm$ 0.1639 & 0.5281 $\pm$ 0.1812      \\ 
\multicolumn{2}{|l|}{EvoGraph ($Y_2$)}         & 0.2181 $\pm$ 0.0170 & 0.6831 $\pm$ 0.1467 & 0.6755 $\pm$ 0.1374 & 0.5274 $\pm$ 0.0582 & 0.4695 $\pm$ 0.1597 & 0.0890 $\pm$ 0.0082       \\ 
\multicolumn{2}{|l|}{GraphRNN-S ($Y_3$)}        & 0.1460 $\pm$ 0.0580 & 0.4338 $\pm$ 0.2094 & 0.4116 $\pm$ 0.2001 & 0.6962 $\pm$ 0.1641 & 0.7666 $\pm$ 0.1146 & 0.0898 $\pm$ 0.0732      \\ 
\multicolumn{2}{|l|}{GraphRNN ($Y_4$)}   & 0.1293 $\pm$ 0.0622 & 0.4620 $\pm$ 0.2304 & 0.4620 $\pm$ 0.2264 & 0.6802 $\pm$ 0.1689 & 0.7600 $\pm$ 0.1219 & 0.0930 $\pm$ 0.0740     \\ 
\multicolumn{2}{|l|}{DeepNC ($Y_5$)}    & 0.4830 $\pm$ 0.0657 & 0.2984 $\pm$ 0.1203 & 0.3049 $\pm$ 0.1365 & 0.5525 $\pm$ 0.2005 & 0.6642 $\pm$ 0.1505 & 0.1355 $\pm$ 0.0177     \\ \hline 
\multicolumn{2}{|l|}{SCGG ($X$)}       & \textbf{0.0701} $\pm$ 0.0232 & \textbf{0.2905} $\pm$ 0.1129 & \textbf{0.2871} $\pm$ 0.1083 & \textbf{0.2046} $\pm$ 0.0553 & \textbf{0.2688} $\pm$ 0.0671 & \textbf{0.0626} $\pm$ 0.0135  \\
\Xhline{6\arrayrulewidth}
\multirow{5}{*}{ Gain} & $\frac{Y_1 - X}{Y_1} \times 100$ & 88.07 & 48.44 & 49.19 & 64.09 & 58.78 & 88.15  \\ 
& $\frac{Y_2 - X}{Y_2} \times 100$ & 67.86 & 57.47 & 57.50 & 61.21 & 42.75 & 29.66       \\ 
& $\frac{Y_3 - X}{Y_3} \times 100$ & 51.99 & 33.03 & 30.25 & 70.61 & 64.94 & 30.29      \\ 
& $\frac{Y_4 - X}{Y_4} \times 100$ & 45.78 & 37.12 & 37.86 & 69.66 & 64.63 & 32.69     \\ 
& $\frac{Y_5 - X}{Y_5} \times 100$ & 85.49 & 2.65 & 5.84 & 62.97 & 59.53 & 53.80 \\  \hline 
\end{tabular}
}}}
\caption{\label{tab:results} \textbf{Comparison of SCGG with its competitors for $\pmb{m = 10}$ in terms of GED (Avg. $\pmb{\pm}$ Std.).}}
\end{center}
\end{adjustwidth}
\end{table}

\textcolor{black}{In this subsection, the experiments conducted to evaluate the performance of our proposed method against the baselines are presented in three parts. In the first part, we set the maximum possible value for the parameter $m$ such that the competing methods can be evaluated on all datasets. Then, we compare the obtained results and report the gain of SCGG over the baselines. In the second part, we discretely change the value of $m$ from the lowest to the highest possible amount in such a way that all datasets can be utilized for model testing. Then we study how the performances of various methods are affected by increasing the value of $m$. Finally, in the third part, we raise $m$ to much higher values and evaluate the efficacy of all approaches on the dataset that offers this possibility.}\par
\textcolor{black}{We first analyze the performance of different methods for the case where $m=10$. The reason for choosing this value for $m$ is that, as outlined in Table \ref{tab:dataset}, the minimum number of nodes among graphs of all datasets is 11. Hence, to construct initial graphs $G_0$, a maximum of 10 nodes can be removed from the original graphs. We report the obtained results in Table \ref{tab:results}, from which it is evident that for all datasets, SCGG is the best performing method in terms of the lowest average GED score. More precisely, SCGG obtains an average gain of 51.74\% over other approaches based on the experiments conducted on all datasets, with the lowest gain value of 2.65\% and the highest gain of 88.15\%. Furthermore, in most cases, the standard deviations of our results are less than those of the baselines.}\par 
\textcolor{black}{Besides, the results of Table \ref{tab:results} reveal that KronEM does not perform well in general, so that, unlike other methods, its average GED has never been lower than 0.52. There can be several reasons for this. First, unlike SCGG, GraphRNN-S, GraphRNN, and to some extent DeepNC, this method is not trained on a dataset of graphs, but rather it processes each graph in the test set separately, i.e., it completes the structure of each partially observed graph based solely on the available part of it. Another reason for the underperformance of KronEM might be due to the fact that the Kronecker graphs model generates graphs with $2^x$ nodes. Therefore, when an initial graph $G_0$ is given to KronEM, it increases the number of its nodes to the nearest power of 2. This can lead to a significant difference between the ground truth and the completed graph regarding the number of nodes, thereby causing the GED score to be raised.}\par
\textcolor{black}{In addition to what we have discussed so far regarding the results in Table \ref{tab:results}, they also indicate that EvoGraph considerably underperforms on the IMDBBINARY and IMDBMULTI datasets. This is because the upscaling process of EvoGraph tends to establish connections with new nodes that have not yet been linked to the graph. In other words, adding new edges is performed with a high priority to connect new nodes to the already generated graph, meaning that setting up more connections between the previously added nodes and the nodes of the initial graph $G_0$ is carried out with a relatively low priority. Thus, it is not surprising that the graphs produced by EvoGraph generally contain fewer edges than the ones belonging to the IMDBBINARY or IMDBMULTI datasets, which according to the statistics listed in Table \ref{tab:dataset}, have low edge sparsity. In light of this, we can expect a decrease in the performance of EvoGraph on these two datasets.} \par
\begin{figure}[!h]
\centering
\includegraphics[width=0.8\textwidth]{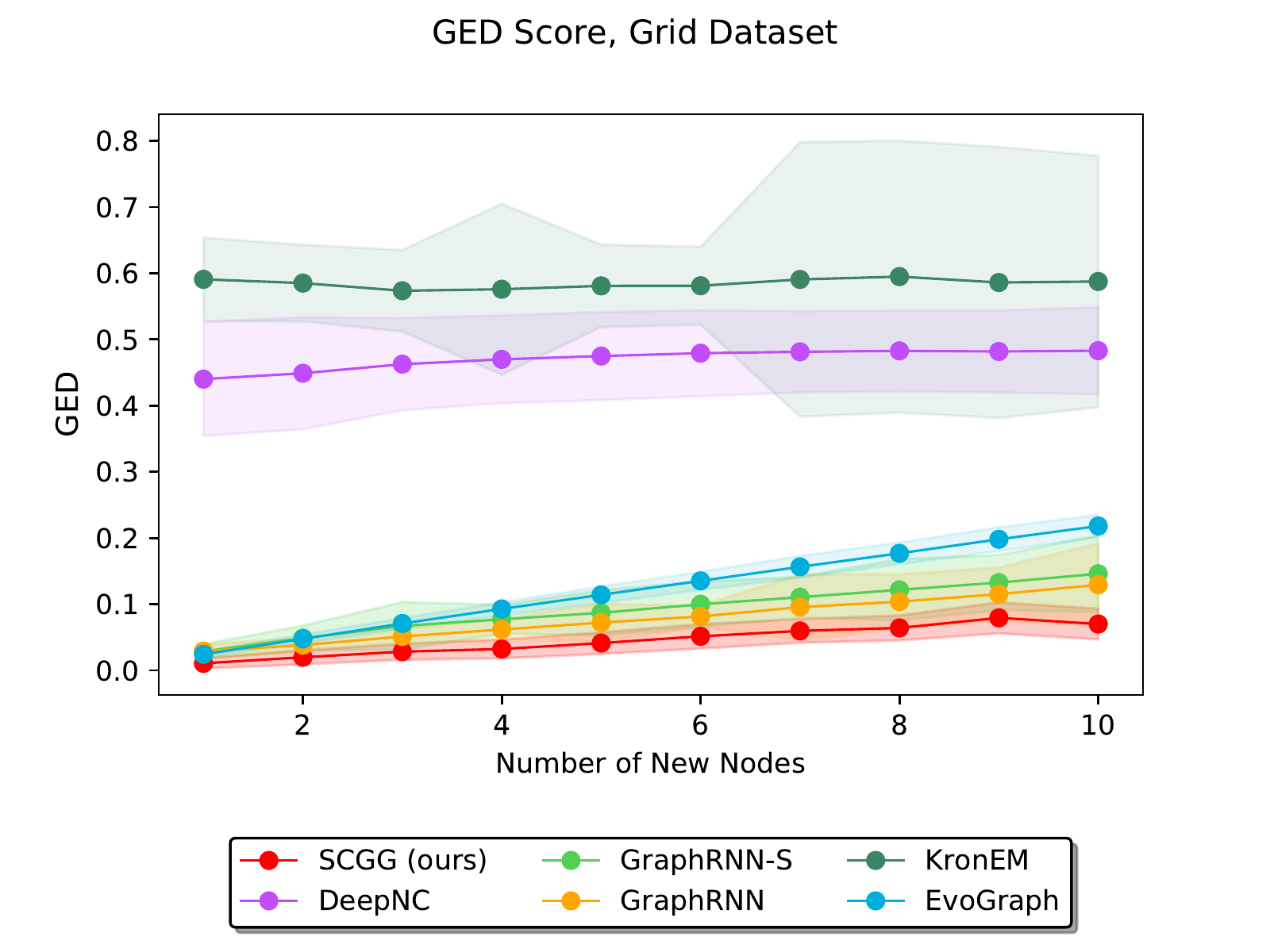}
\vspace*{-1cm}
\begin{center}
\caption{\textcolor{black}{Performance comparison on the Grid dataset in terms of GED (lower is better) as a function of the number of new nodes to be added (i.e., $m$).} }
\label{fig:grid_1_to_10}
\end{center}
\end{figure}
\begin{figure}[!h]
\centering
\includegraphics[width=0.8\textwidth]{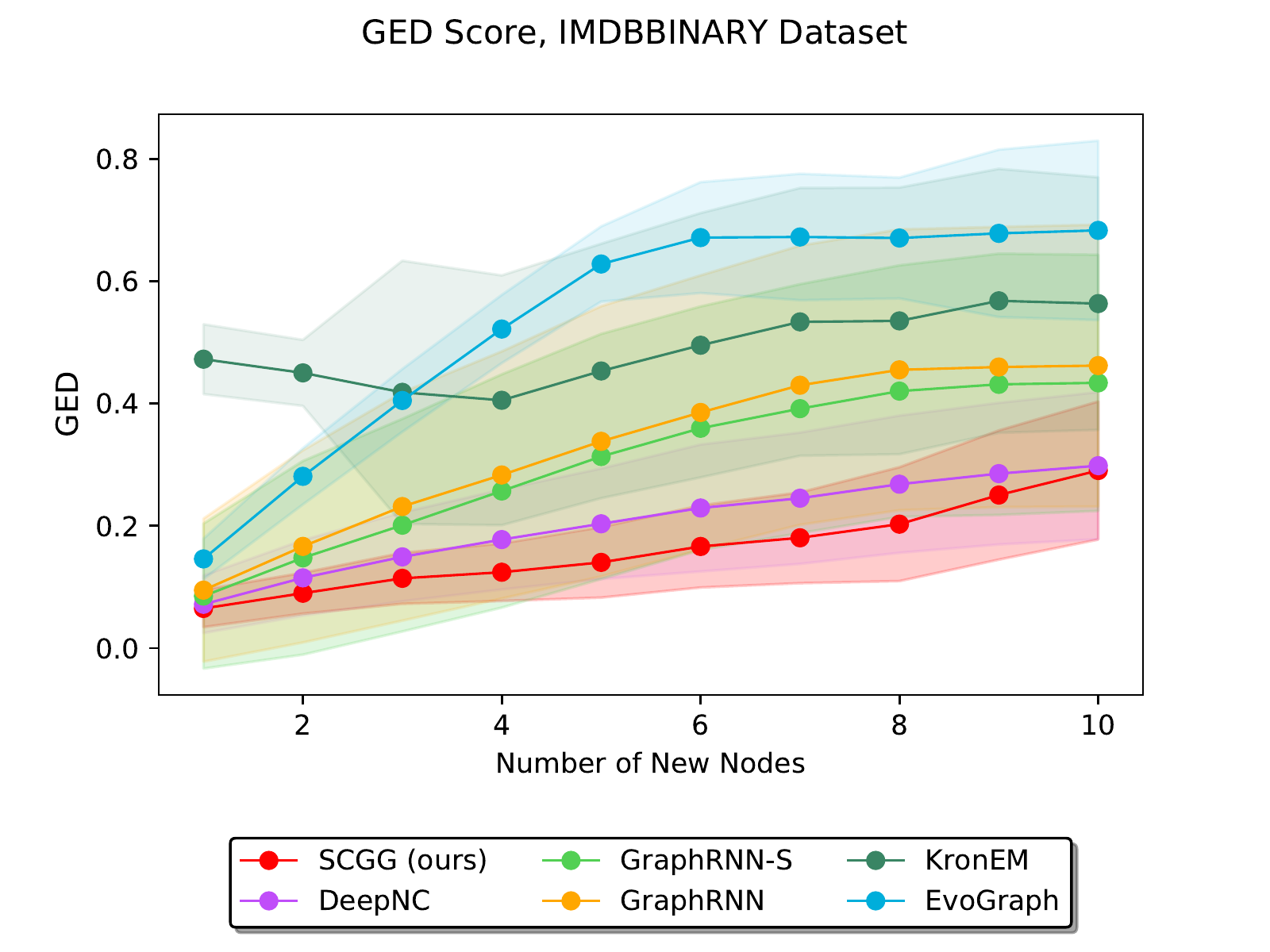}
\vspace*{-1cm}
\begin{center}
\caption{\textcolor{black}{Performance comparison on the IMDBBINARY dataset in terms of GED (lower is better) as a function of the number of new nodes to be added (i.e., $m$).} }
\label{fig:IMDBBINARY_1_to_10}
\end{center}
\end{figure}
\begin{figure}[!h]
\centering
\includegraphics[width=0.8\textwidth]{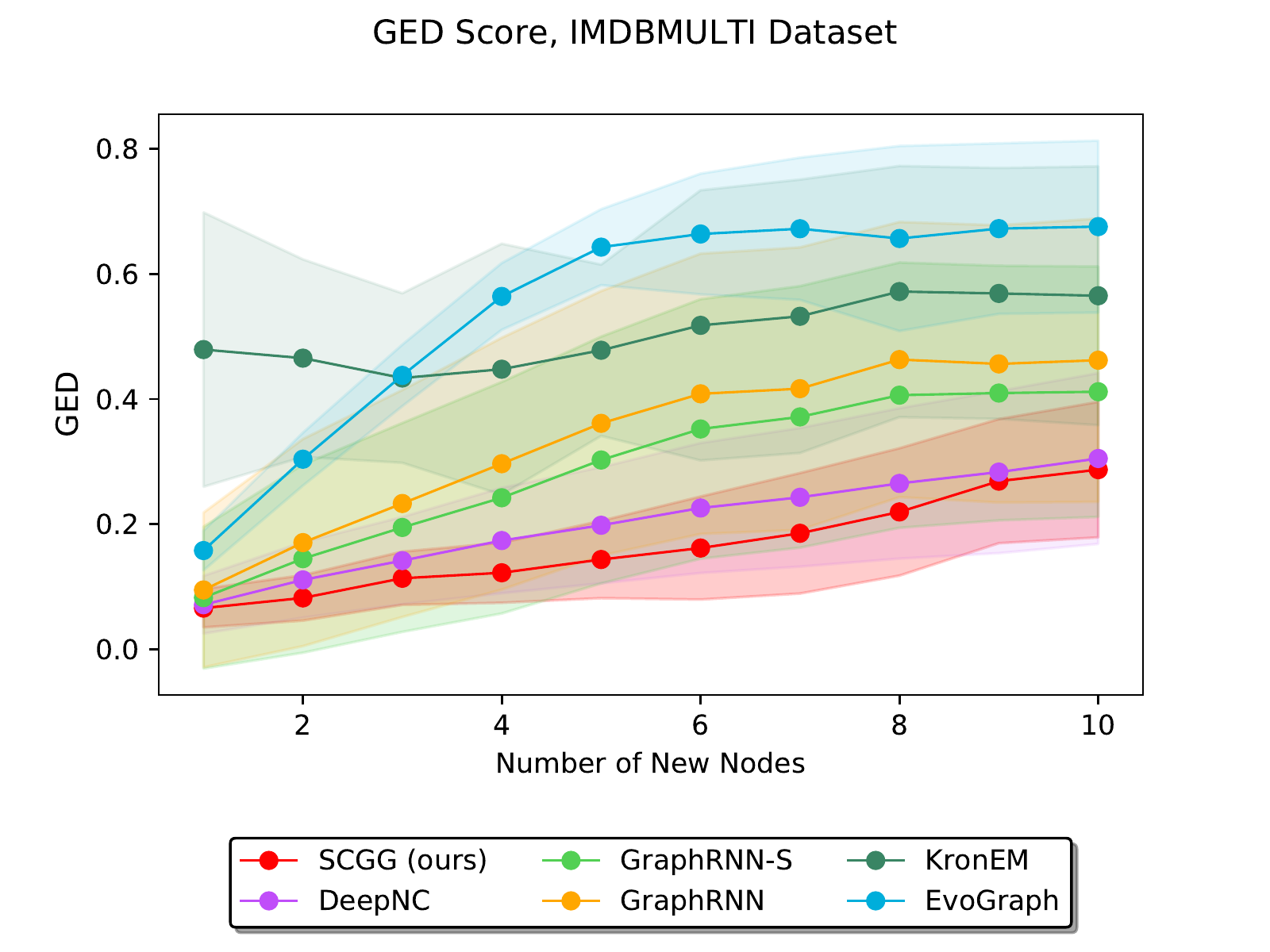}
\vspace*{-1cm}
\begin{center}
\caption{\textcolor{black}{Performance comparison on the IMDBMULTI dataset in terms of GED (lower is better) as a function of the number of new nodes to be added (i.e., $m$).}}
\label{fig:IMDBMULTI_1_to_10}
\end{center}
\end{figure}
\begin{figure}[!h]
\centering
\includegraphics[width=0.8\textwidth]{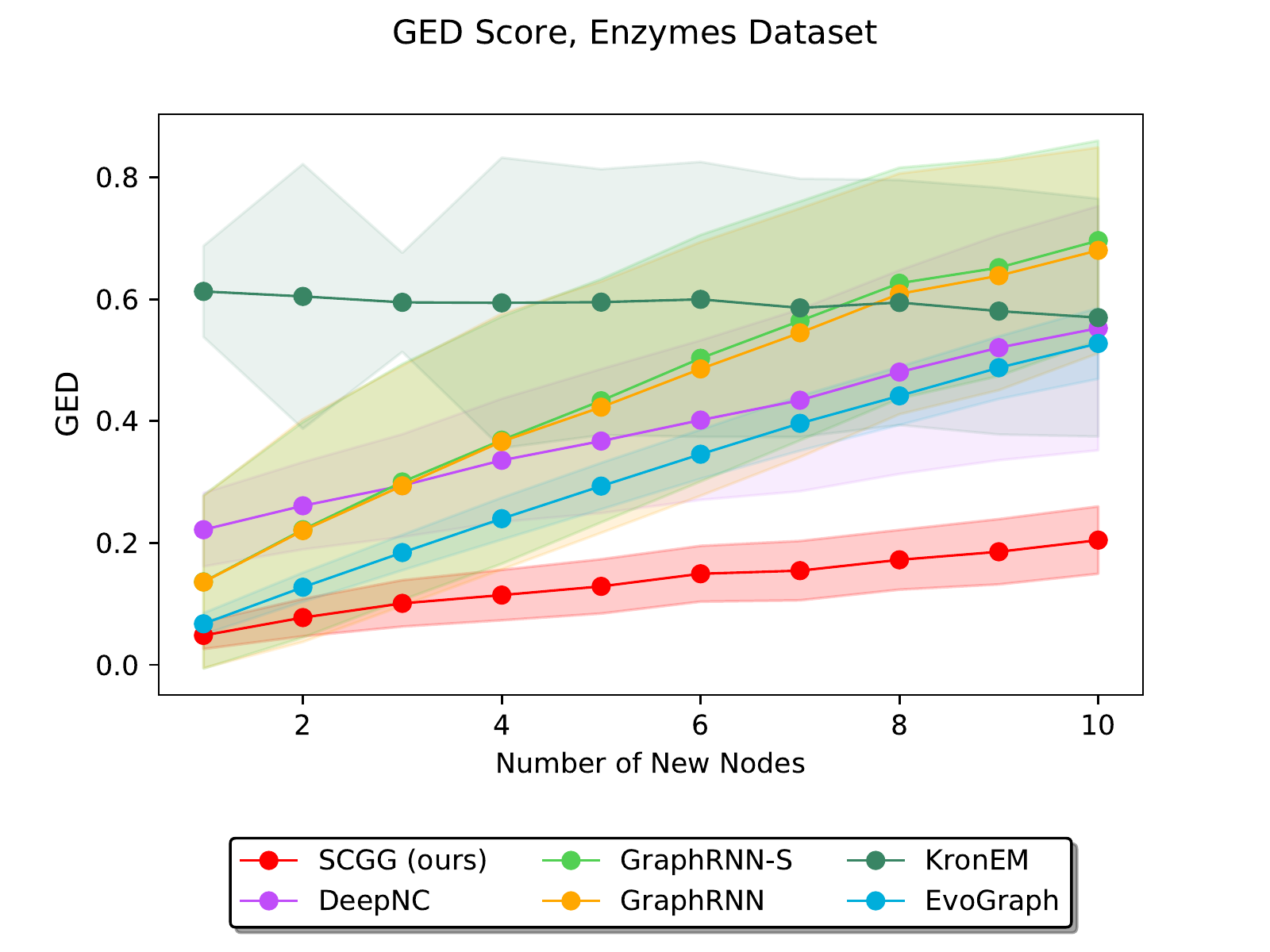}
\vspace*{-1cm}
\begin{center}
\caption{\textcolor{black}{Performance comparison on the Enzymes dataset in terms of GED (lower is better) as a function of the number of new nodes to be added (i.e., $m$).} }
\label{fig:Enzymes_1_to_10}
\end{center}
\end{figure}
\begin{figure}[!h]
\centering
\includegraphics[width=0.8\textwidth]{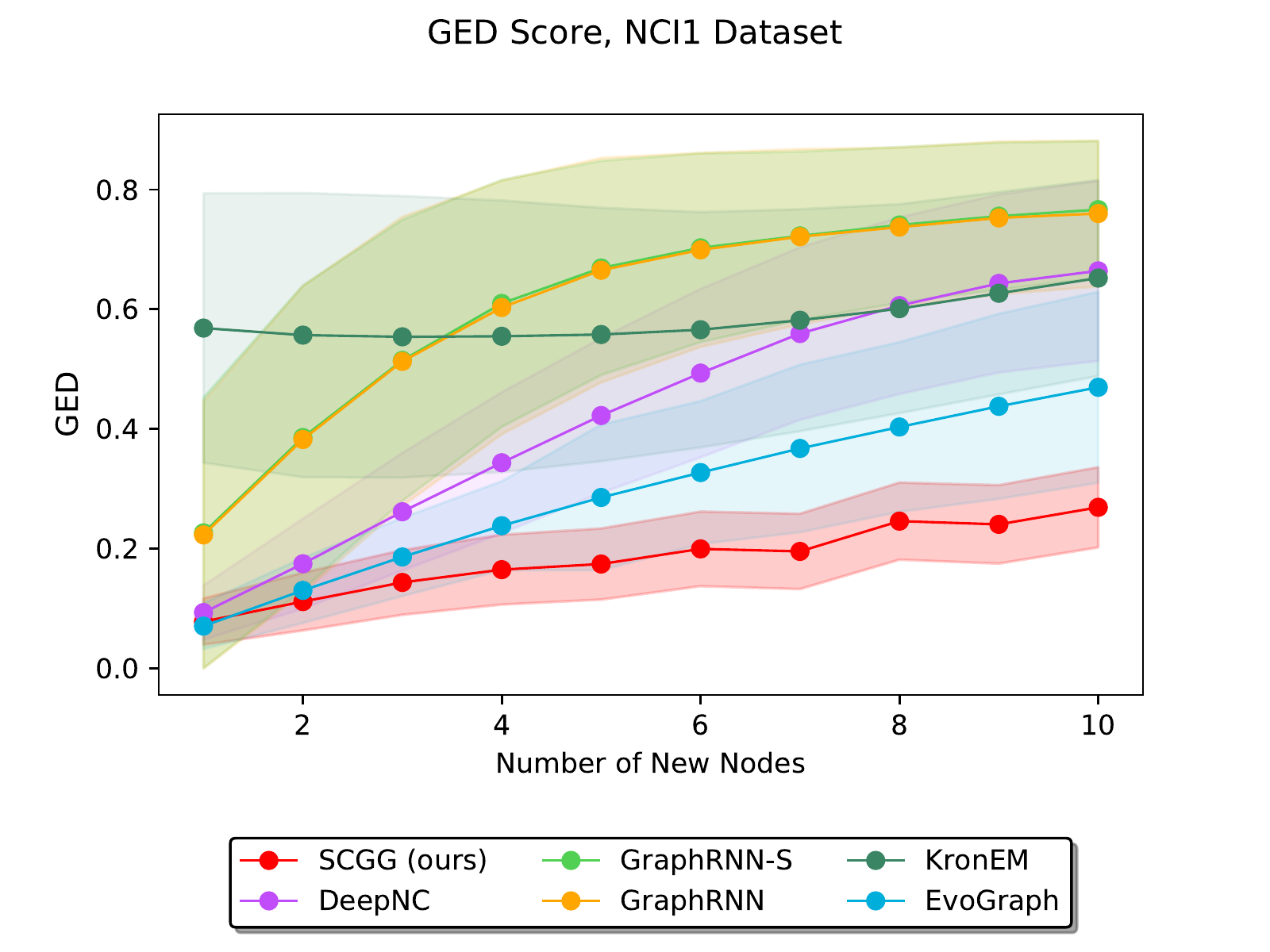}
\vspace*{-1cm}
\begin{center}
\caption{\textcolor{black}{Performance comparison on the NCI1 dataset in terms of GED (lower is better) as a function of the number of new nodes to be added (i.e., $m$).} }
\label{fig:NCI1_1_to_10}
\end{center}
\end{figure}
\begin{figure}[!h]
\centering
\includegraphics[width=0.8\textwidth]{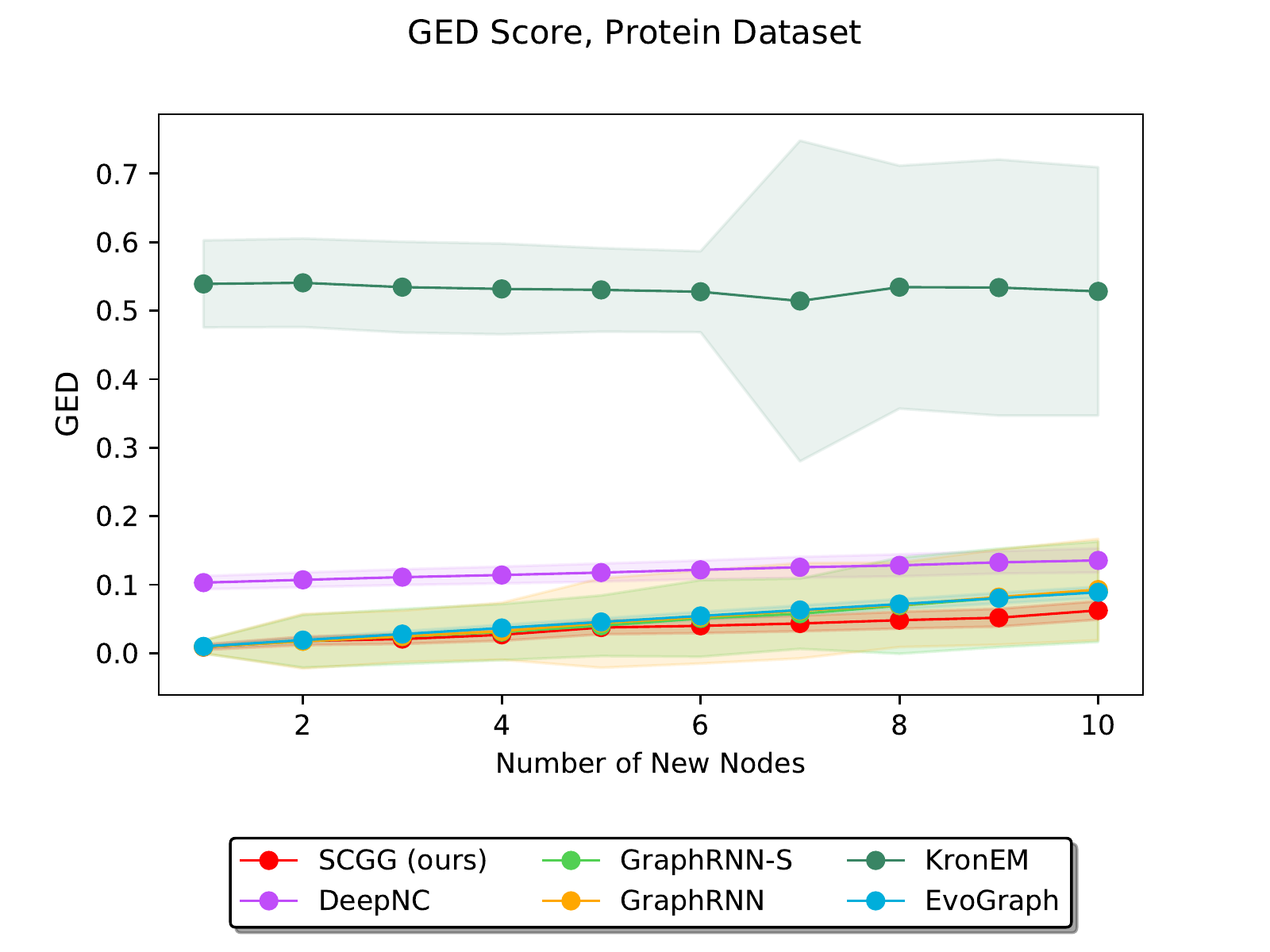}
\vspace*{-1cm}
\begin{center}
\caption{\textcolor{black}{Performance comparison on the Protein dataset in terms of GED (lower is better) as a function of the number of new nodes to be added (i.e., $m$).} }
\label{fig:Protein_1_to_10}
\end{center}
\end{figure}

\textcolor{black}{In the second part of the experiments, we vary the value of $m$ discretely from 1 to 10 and study the performance of different methods as a function of the parameter $m$. In this regard, Figures \ref{fig:grid_1_to_10}, \ref{fig:IMDBBINARY_1_to_10}, \ref{fig:IMDBMULTI_1_to_10}, \ref{fig:Enzymes_1_to_10}, \ref{fig:NCI1_1_to_10}, and \ref{fig:Protein_1_to_10} demonstrate the obtained results on the Grid, IMDBBINARY, IMDBMULTI, Enzymes, NCI1, and Protein datasets, respectively. Moreover, since a part of the results are somewhat visually overlapped, which may affect their readability, we provide the readers with another view of them. In this regard, a pairwise comparison between our SCGG approach and each of the baselines is depicted in a separate subplot for all datasets. Accordingly, the second appearance of the results in Figures \ref{fig:grid_1_to_10}, \ref{fig:IMDBBINARY_1_to_10}, \ref{fig:IMDBMULTI_1_to_10}, \ref{fig:Enzymes_1_to_10}, \ref{fig:NCI1_1_to_10}, and \ref{fig:Protein_1_to_10} can be found in \nameref{S1_Fig}, \nameref{S2_Fig}, \nameref{S3_Fig}, \nameref{S4_Fig}, \nameref{S5_Fig}, and \nameref{S6_Fig}, respectively. In the following, we discuss the results obtained on each dataset.}\par
\textcolor{black}{Fig. \ref{fig:grid_1_to_10} shows the effect of increasing the value of $m$ on the performance of various methods on the Grid dataset. According to these results, the GED values of most methods (i.e., SCGG, GraphRNN-S, GraphRNN, and EvoGraph) increase almost uniformly with the growth of $m$, which makes sense since as $m$ increases, the task becomes more difficult. A noteworthy point here is that our proposed SCGG approach performs the best (lowest GED score). In addition, as $m$ gets higher values, the GED of our approach increases with a lower slope. This figure also demonstrates the poor performance of KronEM (both in terms of the relatively high average GED score and the high standard deviations), which is in accordance with what we discussed before. The results also indicate that DeepNC underperforms on the Grid dataset. This may be due to the fact that DeepNC, unlike other competitors, does not conduct its processing steps by taking into account the whole initial graph $G_0$ at once. To put it another way, other methods receive an initial graph $G_0$ and start adding new nodes on top of it. Meanwhile, DeepNC starts constructing the graph from scratch, and at each stage, it randomly decides whether to choose the next node from the set of initial graph nodes, or add a new one. Therefore, since the graphs of the Grid dataset follow a highly regular structural pattern, not considering whole information of initial graphs at once prior to processing can lead to the performance drop of DeepNC by constructing graphs that are substantially different from the expected ones.}\par
\textcolor{black}{Figures \ref{fig:IMDBBINARY_1_to_10} and \ref{fig:IMDBMULTI_1_to_10} show the results obtained on the IMDBBINARY and IMDBMULTI datasets, respectively. They reveal that for all values of $m$, the SCGG outperforms the baselines. It is also evident from these results that EvoGraph has achieved the worst performance among other competitors. This, as explained earlier, can be due to the tendency of EvoGraph to complete the graph structures by adding a small number of edges to $G_0$, which is in contrast to the non-sparsity of the graphs belonging to these two datasets.}\par

\textcolor{black}{The results on datasets Enzymes and NCI1 are depicted in Figures \ref{fig:Enzymes_1_to_10} and \ref{fig:NCI1_1_to_10}, respectively. Since these two datasets share relatively similar statistical properties, as listed in Table \ref{tab:dataset}, somewhat similar results are observed on them. In this regard, our SCGG approach achieves the best performance compared to other methods. Specifically, in almost all cases it offers the lowest average GED score. Moreover, in the vast majority of circumstances, the standard deviations of the results obtained by our method are lower compared to the other approaches. These results also demonstrate that GraphRNN-S and GraphRNN perform the worst as the value of $m$ increases. This is because these two are general graph generation approaches, which are not specifically designed to solve problems such as structure-conditioned graph generation or graph completion. Therefore, although they have achieved acceptable performance in some cases, it is not surprising that in some other cases they perform poorly compared to the baselines.}\par

\textcolor{black}{Finally, Fig. \ref{fig:Protein_1_to_10} depicts the results on the Protein dataset, in which the value of $m$ varies discretely from 1 to 10. The results indicate that the SCGG method obtains a lower GED than the baseline methods in almost all cases, and as the value of $m$ goes up, this performance superiority more clearly manifests itself. In addition, the weak performance of KronEM can be evidently seen in these results, the reasons for which have been discussed in detail previously.}\par
\begin{figure}[!h]
\centering
\includegraphics[width=0.8\textwidth]{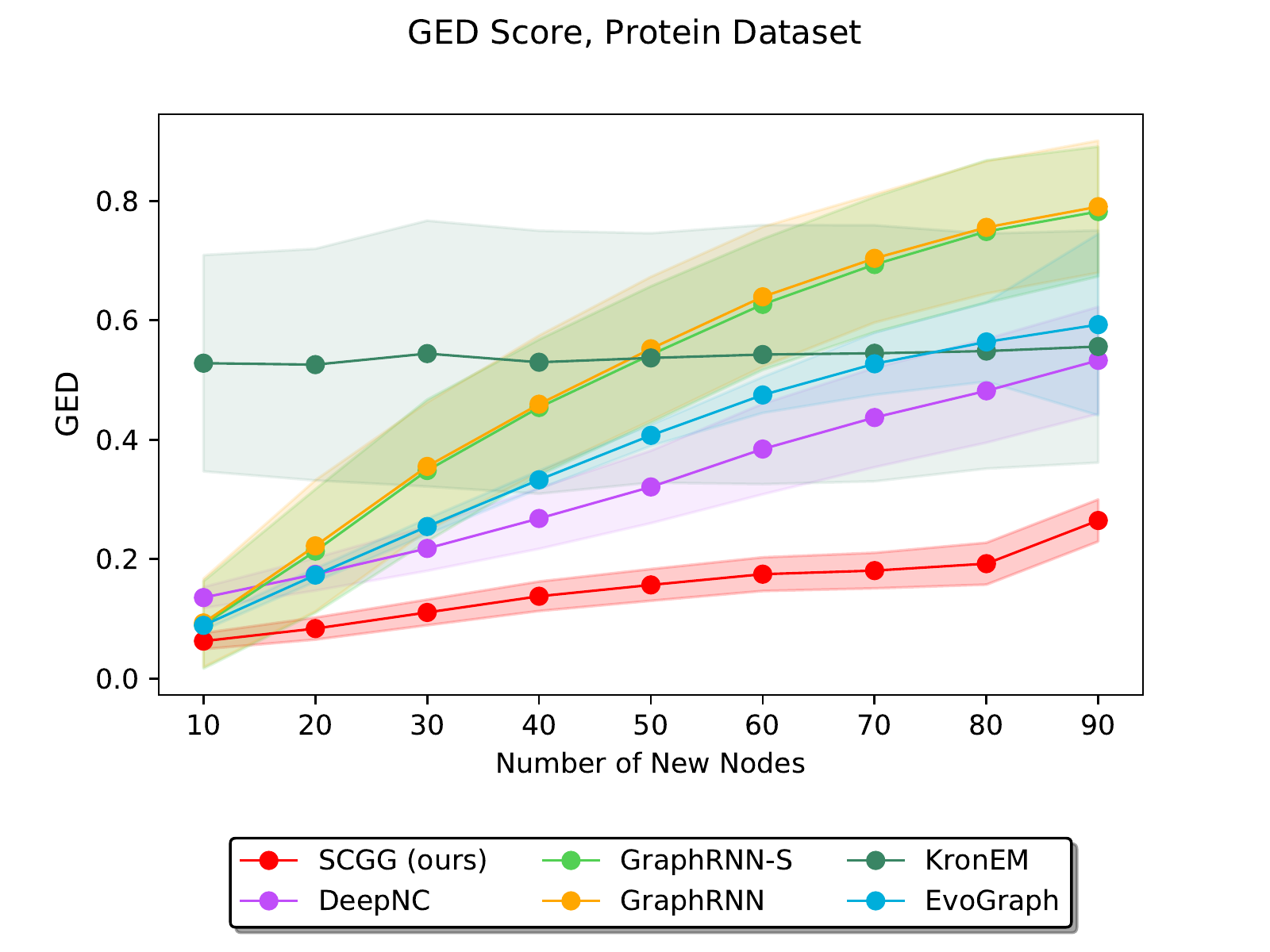}
\vspace*{-1cm}
\begin{center}
\caption{\textcolor{black}{Performance comparison on the Protein dataset in terms of GED (lower is better) as a function of the parameter $m$, which varies discretely from 10 to 90 in steps of 10.}}
\label{fig:Protein_10_to_90}
\end{center}
\end{figure}

\textcolor{black}{In the third part of the experiments, we study the performance of all competing approaches in the case where a much larger number of nodes are supposed to be added to initial graphs $G_0$. Accordingly, we conduct the experiments on the Protein dataset, which, due to the large size of its graphs, gives us this opportunity. More precisely, we increase the value of the parameter $m$ from 10 to 90 (i.e., the maximum possible value that does not exceed the minimum number of nodes in this dataset) in steps of 10. The results of these experiments are illustrated in Fig. \ref{fig:Protein_10_to_90}. As we can see, our method achieves the best results in terms of the lowest GED score for all values of $m$. Furthermore, in the majority of cases and especially as $m$ gets higher values, our results show smaller standard deviations than those of other approaches. We can also observe that for the higher values of $m$, for which both the tasks of graph completion and structure-conditioned graph generation become much more challenging, the performance of GraphRNN-S, GraphRNN, and EvoGraph deteriorate rapidly. This can be interpreted according to the fact that these approaches are not particularly designed to address such tasks. Conversely, as the parameter $m$ rises to its highest values, SCGG, DeepNC, and KronEM offer the best results, respectively.}\par
\textcolor{black}{Another perspective of the results in Fig. \ref{fig:Protein_10_to_90} can be found in \nameref{S7_Fig}, providing the readers with a pairwise comparison of our SCGG model and each of the baselines.} 

\section{Conclusions}\label{sec:conclusion}
\textcolor{black}{In this work, we have presented SCGG, a novel structure-conditioned graph generation approach that autoregressively generates a graph by adding new nodes and their corresponding edges on top of a given initial substructure $G_0$. Specifically, the architecture of our model consists of a specific graph representation learning network, which is the main responsible for considering the conditioning substructure, and an autoregressive generative model (i.e., a recurrent neural network) that mostly maintains the generation history. We then have employed this model to address the intrinsically hard-to-solve problem of network completion, in which the goal is to complete the structure of a partially observed graph, some of whose nodes are totally unknown. To demonstrate the superiority of our proposed SCGG model, we have conducted intensive experiments on both synthetic and real-world datasets and compared the performance of our method against state-of-the-art baselines for the task of graph completion. The experimental results illustrate that SCGG outperforms the baselines in terms of the GED score, which indicates that the graphs generated by our model, on average, are the closest to the ground truth graphs. To the best of our knowledge, this is the first time a completely deep learning-based approach addresses the graph completion problem.}\par
\textcolor{black}{Potential research pathways to be explored in the future include extending the SCGG model in such a way that it can be used for molecular graph generation, in which the existence of predetermined chemical substructures in the final designed molecules confers specific chemical properties to them. Furthermore, another future research direction is to enhance model scalability, so that the SCGG can generate even much larger graphs.}

\section*{Supporting information}

\paragraph*{S1 Fig.}
\label{S1_Fig}
{\bf \textcolor{black}{Pairwise performance comparison between our proposed SCGG method and its competitors on the Grid dataset.}} \textcolor{black}{The results are reported in terms of GED (the lower the better) as a function of the number of new nodes (denoted by $m$) that are added to initial graphs (each represented by the notation $G_0$ in the paper).}
\begin{figure}[!h]
\centering
\includegraphics[width=0.95\textwidth]{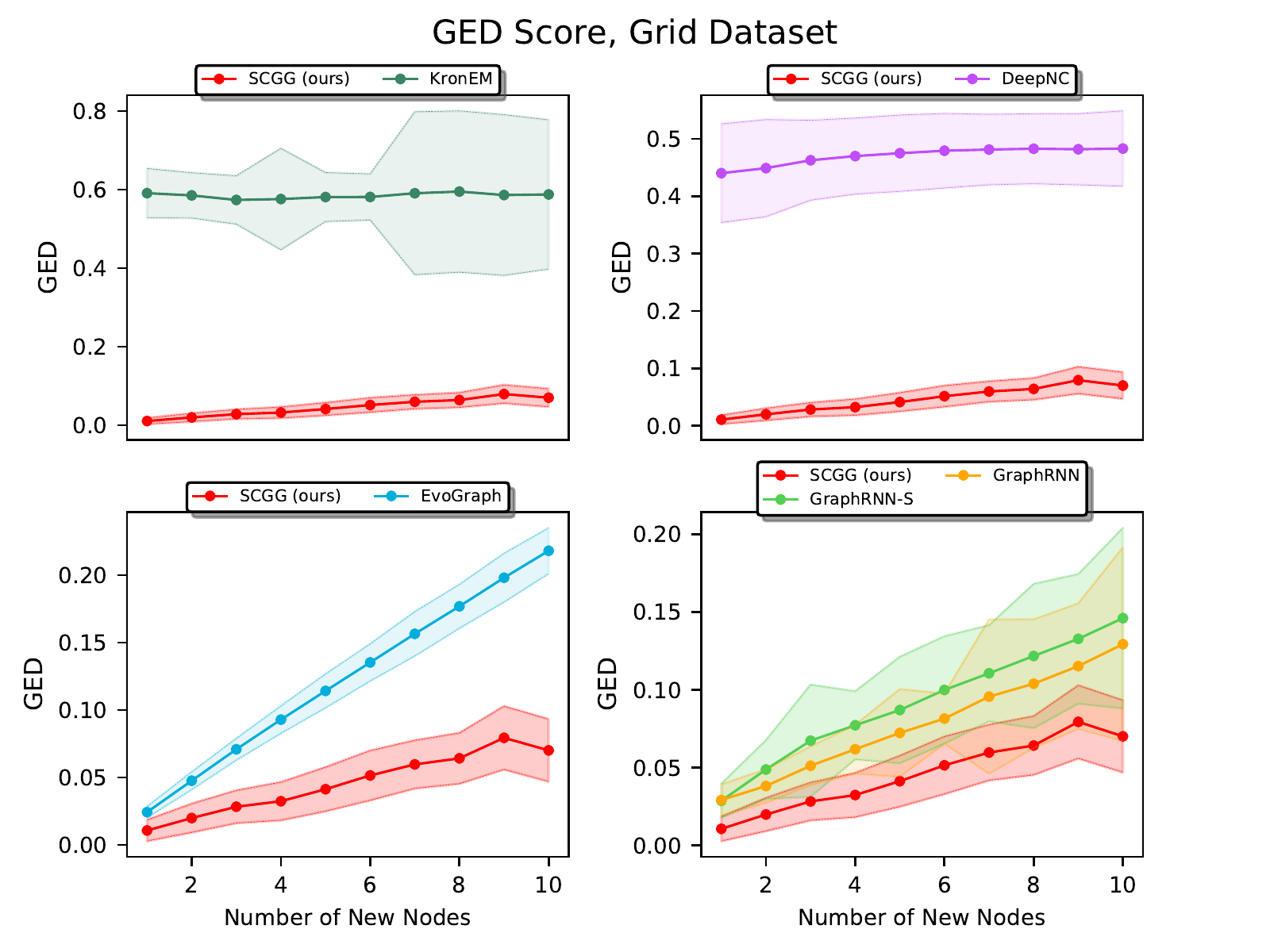}
\vspace*{-1cm}
\begin{center}
\end{center}
\end{figure}
\newpage
\paragraph*{S2 Fig.}
\label{S2_Fig}
{\bf \textcolor{black}{Pairwise performance comparison between our proposed SCGG method and its competitors on the IMDBBINARY dataset.}} \textcolor{black}{The results are reported in terms of GED (the lower the better) as a function of the number of new nodes (denoted by $m$) that are added to initial graphs (each represented by the notation $G_0$ in the paper).}
\begin{figure}[!h]
\centering
\includegraphics[width=0.95\textwidth]{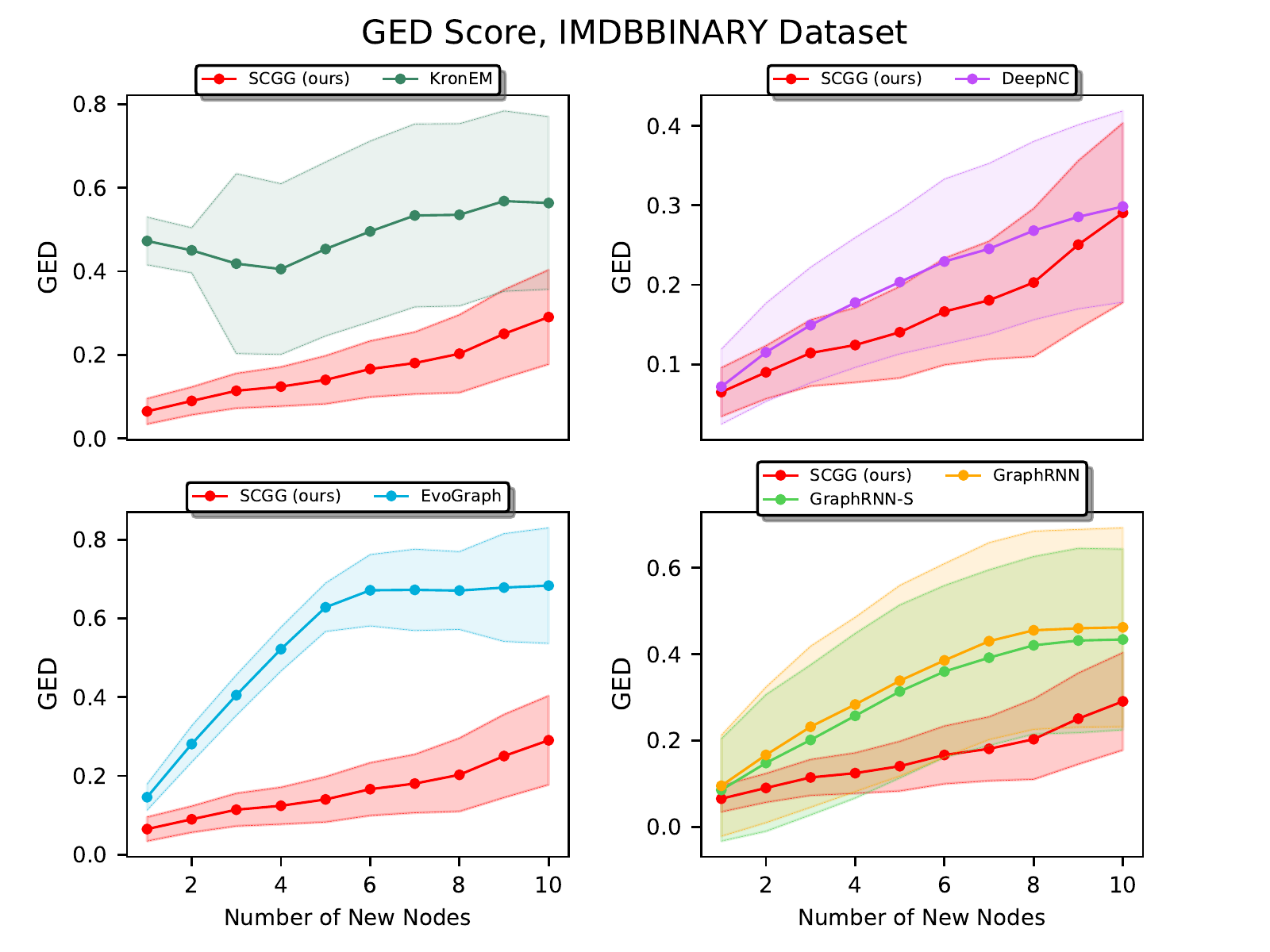}
\vspace*{-1cm}
\begin{center}
\end{center}
\end{figure}
\newpage
\paragraph*{S3 Fig.}
\label{S3_Fig}
{\bf \textcolor{black}{Pairwise performance comparison between our proposed SCGG method and its competitors on the IMDBMULTI dataset.}} \textcolor{black}{The results are reported in terms of GED (the lower the better) as a function of the number of new nodes (denoted by $m$) that are added to initial graphs (each represented by the notation $G_0$ in the paper).}
\begin{figure}[!h]
\centering
\includegraphics[width=0.95\textwidth]{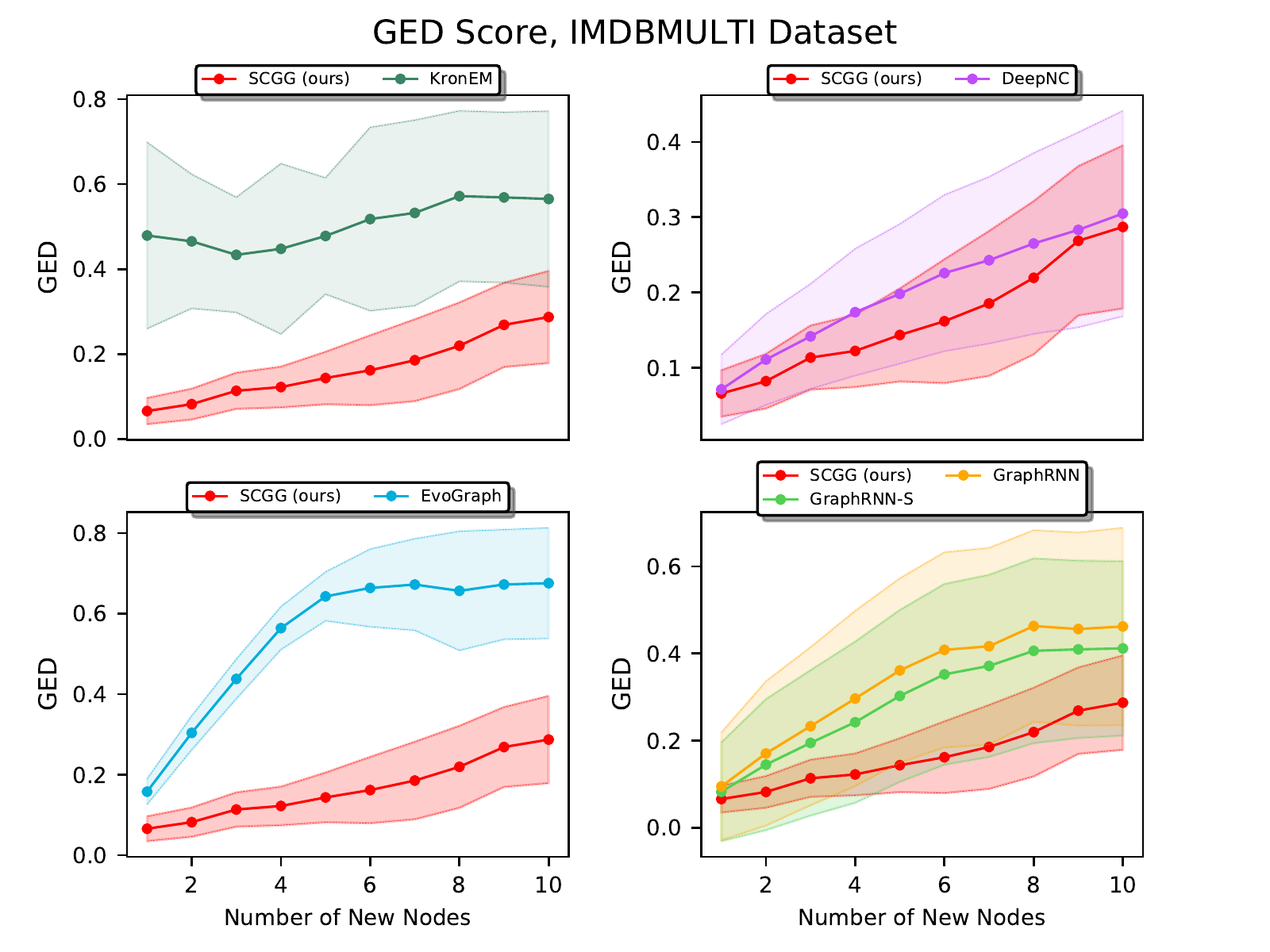}
\vspace*{-1cm}
\begin{center}
\end{center}
\end{figure}
\newpage
\paragraph*{S4 Fig.}
\label{S4_Fig}
{\bf \textcolor{black}{Pairwise performance comparison between our proposed SCGG method and its competitors on the Enzymes dataset.}} \textcolor{black}{The results are reported in terms of GED (the lower the better) as a function of the number of new nodes (denoted by $m$) that are added to initial graphs (each represented by the notation $G_0$ in the paper).}
\begin{figure}[!h]
\centering
\includegraphics[width=0.95\textwidth]{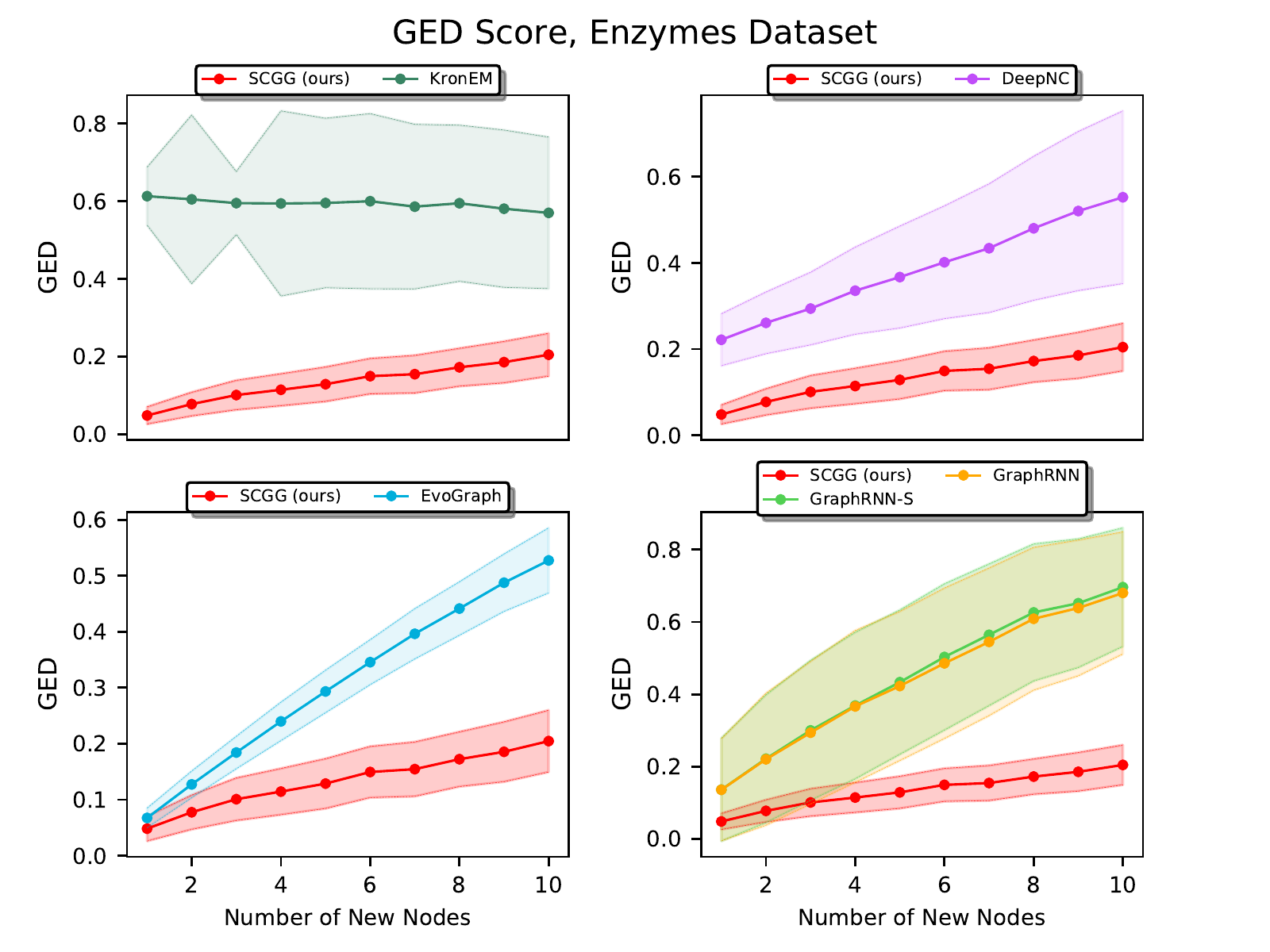}
\vspace*{-1cm}
\begin{center}
\end{center}
\end{figure}
\newpage
\paragraph*{S5 Fig.}
\label{S5_Fig}
{\bf \textcolor{black}{Pairwise performance comparison between our proposed SCGG method and its competitors on the NCI1 dataset.}} \textcolor{black}{The results are reported in terms of GED (the lower the better) as a function of the number of new nodes (denoted by $m$) that are added to initial graphs (each represented by the notation $G_0$ in the paper).}
\begin{figure}[!h]
\centering
\includegraphics[width=0.95\textwidth]{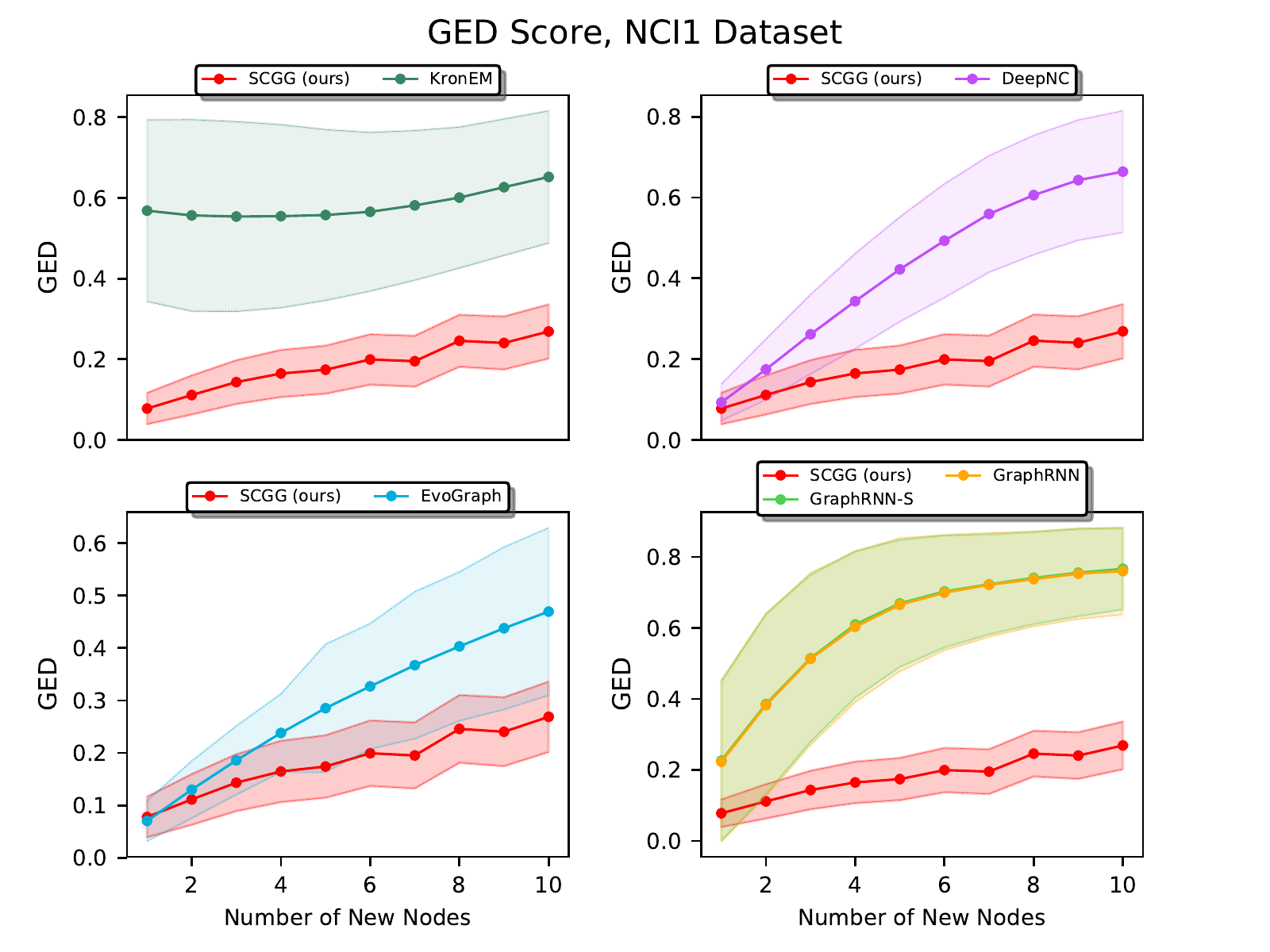}
\vspace*{-1cm}
\begin{center}
\end{center}
\end{figure}
\newpage
\paragraph*{S6 Fig.}
\label{S6_Fig}
{\bf \textcolor{black}{Pairwise performance comparison between our proposed SCGG method and its competitors on the Protein dataset.}} \textcolor{black}{The results are reported in terms of GED (the lower the better) as a function of the number of new nodes (denoted by $m$) that are added to initial graphs (each represented by the notation $G_0$ in the paper).}
\begin{figure}[!h]
\centering
\includegraphics[width=0.95\textwidth]{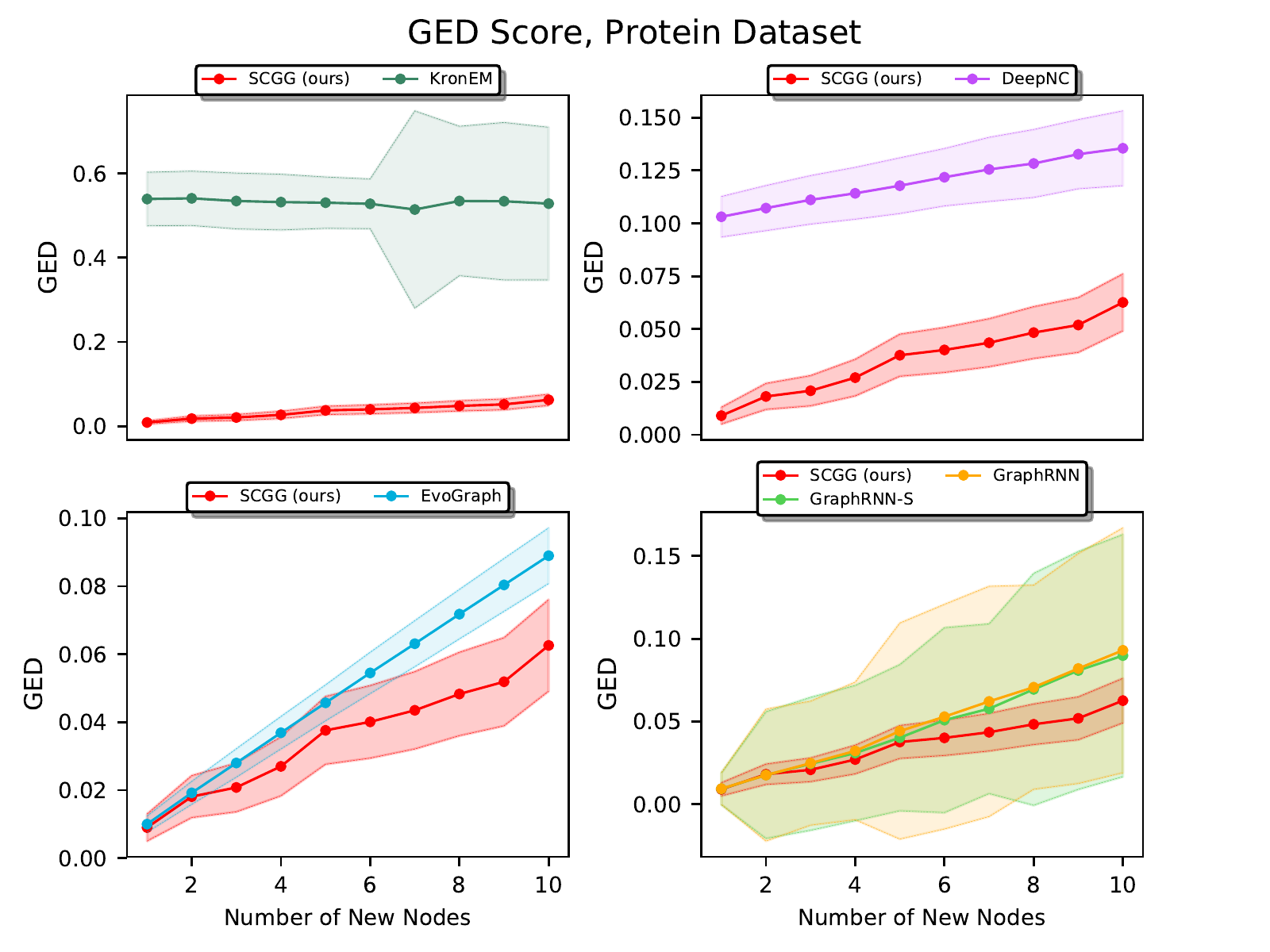}
\vspace*{-1cm}
\begin{center}
\end{center}
\end{figure}
\newpage
\paragraph*{S7 Fig.}
\label{S7_Fig}
{\bf \textcolor{black}{Pairwise performance comparison between our proposed SCGG method and its competitors on the Protein dataset.}} \textcolor{black}{The results are reported in terms of GED (the lower the better) as a function of the parameter $m$ that increases discretely from 10 to 90 in steps of 10.}
\begin{figure}[!h]
\centering
\includegraphics[width=0.95\textwidth]{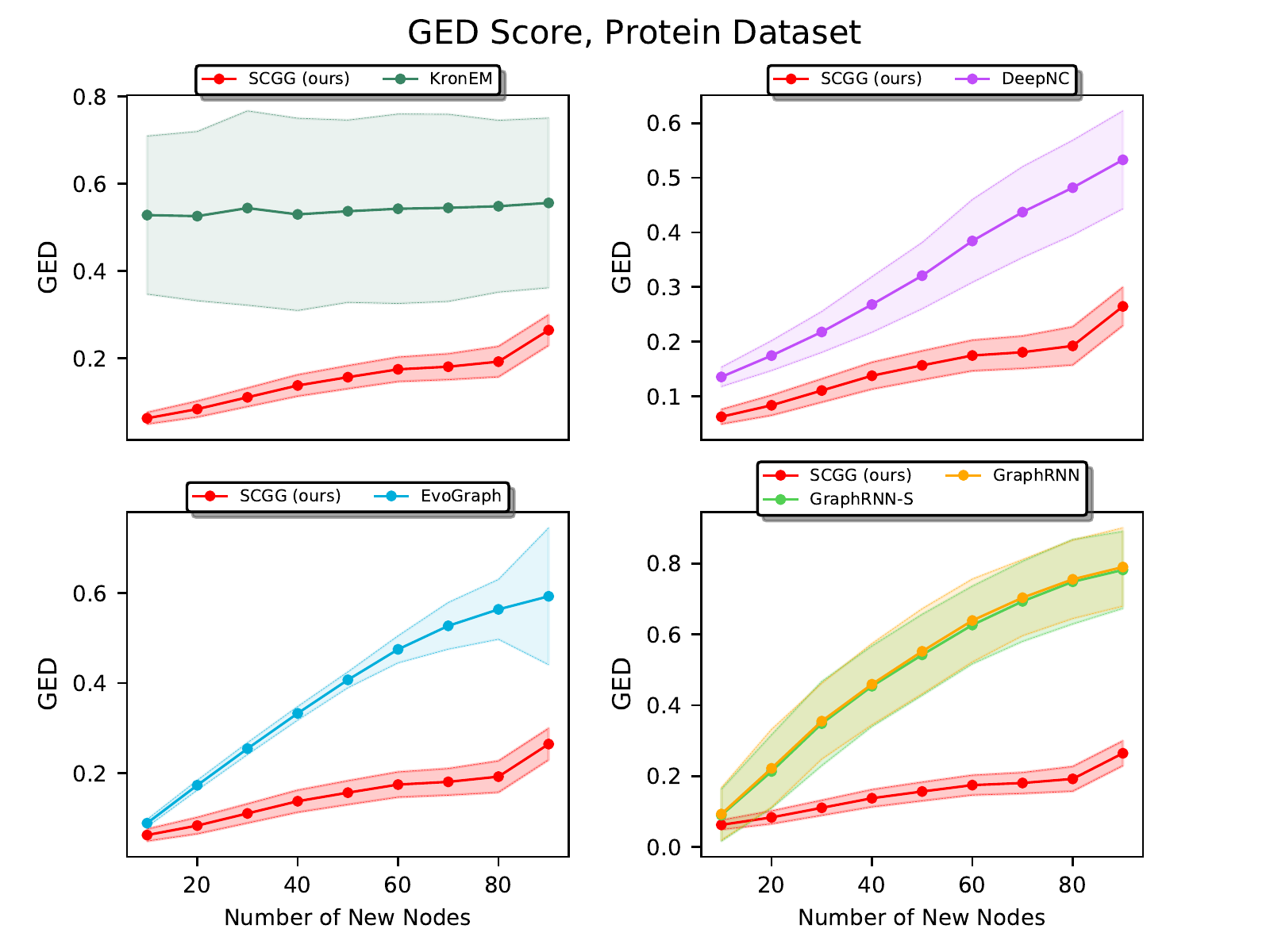}
\vspace*{-1cm}
\begin{center}
\end{center}
\end{figure}

\nolinenumbers

%
%
%

\end{document}